%% file: main-arxiv.tex
\documentclass{article}

\usepackage{iclr2025_conference}

\usepackage[utf8]{inputenc} % allow utf-8 input
\usepackage[T1]{fontenc}    % use 8-bit T1 fonts
\usepackage{hyperref}       % hyperlinks
\usepackage{url}            % simple URL typesetting
\usepackage{booktabs}       % professional-quality tables
\usepackage{nicefrac}       % compact symbols for 1/2, etc.
\usepackage{microtype}      % microtypography
\usepackage{xcolor}         % colors
\usepackage{tikz} 
\usepackage{tikz-3dplot}
\usepackage{wrapfig}
\usepackage{bm}
\usepackage{subcaption}
\usepackage{enumerate}
\usepackage[shortlabels]{enumitem}
\usepackage{pgfplots}
\pgfplotsset{compat=newest}
\usetikzlibrary{positioning}
\usetikzlibrary{shapes,arrows,chains}

\newcommand{\myparagraph}[1]{\noindent\textbf{#1}}
\input{math_commands}

\title{A Theoretical Study of Neural Network Expressive Power via Manifold Topology}
\author{Jiachen Yao$^1$, Mayank Goswami$^2$, Chao Chen$^1$ \\
$^1$Stony Brook University, $^2$CUNY Queens College \\
$^1$\texttt{\{jiachen.yao,chao.chen.1\}@stonybrook.edu} \\ $^2$\texttt{mayank.isi@gmail.com}
}

\date{}

\iclrfinalcopy % Uncomment for camera-ready version, but NOT for submission.

\begin{document}

\maketitle

\begin{abstract}
   A prevalent assumption regarding real-world data is that it lies on or close to a low-dimensional manifold. When deploying a neural network on data manifolds, the required size, i.e., the number of neurons of the network, heavily depends on the intricacy of the underlying latent manifold. While significant advancements have been made in understanding the geometric attributes of manifolds, it's essential to recognize that topology, too, is a fundamental characteristic of manifolds. In this study, we investigate network expressive power in terms of the latent data manifold. Integrating both topological and geometric facets of the data manifold, we present a size upper bound of ReLU neural networks. 
   % This bound integrates both topological and geometric facets of manifolds.
   %, with empirical evidence confirming the tightness. 
   %To our knowledge, this constitutes the pioneering effort in investigating the relationship between network size and manifold topology within the context of manifold learning.
   % A prevalent assumption regarding real-world data is that it lies on or close to a lower-dimensional manifold. Machine learning methods that operate under this premise are commonly referred to as manifold learning. When deploying a neural network for manifold learning, the required size of the network heavily depends on the intricacy of the underlying latent manifolds. While significant advancements have been made in understanding the geometric attributes of manifolds, it's essential to recognize that topology, too, is a fundamental characteristic of manifolds. In this study, we delve into a manifold classification challenge and present an upper bound on the size of ReLU neural networks. This bound integrates the topological facets of manifolds,  with empirical evidence confirming the tightness. To our knowledge, this constitutes the pioneering effort in investigating the relationship between network size and manifold topology within the context of manifold learning.
\end{abstract}

\section{Introduction}
\label{sec:intro}
\input{introduction}

\section{Preliminaries}
\label{sec:pre}
\input{preliminary}

\section{Main Results}
\label{sec:complexity}
\input{complexity}

% \section{Empirical Validation}
% \label{sec:experiment}
% \input{experiment}

\section{Conclusion}
\label{sec:discussion}
\input{discussion}

%\bibliography{reference}
\bibliographystyle{iclr2025_conference}

\newpage
\appendix
\section{Appendix}
\input{appendix}

\end{document}

%% file: math_commands.tex
\usepackage{amsmath,amsfonts,bm}
\usepackage{amsthm}

\newtheorem{theorem}{Theorem}

\newtheorem{definition}{Definition}
\newtheorem{lemma}{Lemma}
\newtheorem{proposition}{Proposition}

\newtheorem{claim}{Claim}
\newtheorem*{main*}{Main Theorem}

\def\Figref#1{Figure~\ref{#1}}

\def\secref#1{section~\ref{#1}}
\def\eqref#1{equation~\ref{#1}}
\def\1{\bm{1}}

\def\rvc{{\mathbf{c}}}

\def\rvp{{\mathbf{p}}}
\def\rvq{{\mathbf{q}}}

\def\rvx{{\mathbf{x}}}
\def\rvy{{\mathbf{y}}}

\DeclareMathAlphabet{\mathsfit}{\encodingdefault}{\sfdefault}{m}{sl}
\SetMathAlphabet{\mathsfit}{bold}{\encodingdefault}{\sfdefault}{bx}{n}

\def\gB{{\mathcal{B}}}

\def\gM{{\mathcal{M}}}
\def\gN{{\mathcal{N}}}

\def\gT{{\mathcal{T}}}

\def\sM{{\mathbb{M}}}

\def\sR{{\mathbb{R}}}

\newcommand{\R}{\mathbb{R}}

%% file: introduction.tex
The expressive power of deep neural networks (DNNs) is believed to play a critical role in their astonishing performance. Despite a rapidly expanding literature, the theoretical understanding of such expressive power remains limited. The well-known \textit{universal approximation theorems} \citep{hornik1989multilayer, cybenko1989approximation, leshno1993multilayer, hanin2017approximating} guarantee that neural networks can approximate vast families of functions with an arbitrarily high accuracy. However, the theoretical upper bound of the size of such networks is rather pessimistic; it is exponential to the input space dimension. Indeed, these bounds tend to be loose, because the analyses are often oblivious to the intrinsic structure of the data. Real-world data such as images are believed to live in a manifold of a much lower dimension \citep{lleRoweis2000,tsneVandermaaten2008,pcaJolliffe2016}. 
% A common belief is that real-world data sets can be modeled as samples lying on or close to a compact lower-dimensional manifold \citep{tsneVandermaaten2008, pcaJolliffe2016}. 
%Machine learning approaches that capitalize on this perspective are collectively termed `manifold learning'. 
Such manifold's structure can be used to achieve better bounds of network size.
It has been shown that the network size can be bounded by exponential of the manifold's intrinsic dimension rather than the encompassing input space dimension \citep{lowmanifoldChen2019, schmidt2019deep}.

However, the intrinsic dimension is only a small part of the manifold's property.
% Yet, manifold properties other than the dimension are still mystery towards network size, i.e., the depth and size of neural networks. 
It is natural to ask whether other properties of the manifold, such as topology and geometry, may lead to improved bounds. 
\cite{Safran2016DepthWidthTI} demonstrate that to approximate the indicator function of a $d$-dimensional ball, one only needs a network of size quadratic to $d$. However, this work assumes a rather simplistic input. To extend to a more general setting, one needs to incorporate the topology and geometry of the manifold into the analysis.

Early research has probed the geometry and topology of manifolds. Notably, \cite{federer1959curvature, amenta1998surface} introduce a pivotal curvature measure, which adeptly captures the global geometric nuances of manifolds and has been embraced in manifold learning studies \citep{narayanan2009sample, samplecomplexNaray2010,labeledcomplex2019ramamurthy}. 
%\cc{I think no need to define reach here. Just say these works use geometry to characterize manifold. Later when you introduce condition number, you should mention reach and cite these papers.}
On the topological front, descriptors like \emph{Betti numbers} have been formalized in the language of algebraic topology to characterized the numbers of connected components and holes of a manifold \citep{hatcher2002algebraic, bott1982differential,munkres2018elements, rieckneural}. In their seminal work, \cite{Niyogi2008Homology} integrate manifold's geometry and topology, setting forth conditions for topologically faithful reconstructions grounded in geometric metrics. 
% With the advent of the deep learning era, there has been a burgeoning interest in discerning the interplay between network size and manifold's intrinsic structural attributes. 
%Beyond the previously mentioned studies on function approximation, where functions are characterized over manifolds,  in diverse deep learning applications. \cc{The last sentence is distracting. Maybe put it in the related work?}

With the advent of the deep learning era, there has been a burgeoning interest in discerning the interplay between network size and manifold's intrinsic structural attributes. 
Existing studies \citep{lsh2021dikkala,chartae2019stefan} bound network size with the geometry of manifolds. However, a theoretical framework that successfully integrates network size and topological traits has not yet been developed. This is a missed opportunity. The topological complexity of manifolds plays a crucial role in the learning problem, particularly concerning network size. Empirical findings \citep{guss2018characterizing,naitzat2020topology} suggest that even with similar geometry, data with larger topological complexity requires a larger network. These empirical observations highlight the need for a theoretical analysis that examines how manifold topology and geometry interact with network size. Addressing this need poses a significant challenge, as incorporating topological descriptors into the current analytical framework is inherently difficult due to the discrete nature of topology.

In this paper, we address this gap by presenting an innovative theoretical framework that integrates topology with neural network size. We tackle the challenge by first approximating a homeomorphism to disentangle the geometry and topology of a manifold. Notably, manifold topology remains invariant under a homeomorphism. This allows us to simplify the geometric complexity while maintaining the topological complexity. We then successfully represent discrete topological complexity as a combination of the complexities of basic topological shapes, thus overcoming the challenge posed by the discrete nature of topological descriptors. The resulting upper bound is obtained by constructing a neural network that first learns a low-dimensional embedding of the input manifold, followed by classification in the embedding space. This approach aligns with modern neural network design principles.
%In this paper, we explore how the topology and geometry of manifolds influence network size in classification scenarios. Our results, derived methodically through construction, follow two steps.First, we approximate a homeomorphism between the input manifold and a latent one; second, we carry out classification within this latent manifold. This latent manifold is designed to have simple geometric features, akin to those found in spheres and tori, while retaining the intrinsic topological characteristics of the original manifold. By design, the first phase is purely geometric, as the topological traits remain unaltered, while the subsequent classification phase is predominantly topological. Consequently, the required network size can be delineated into two distinct parts. We employ \textit{Betti numbers} and the \textit{condition number} as metrics to gauge topological and geometric complexities, respectively. Specifically, Betti numbers quantify the number of connected components and holes within the manifold, whereas the condition number characterizes the manifold's overall curvature.
%We propose a theoretical framework to independently analyze the influences of topology and geometry on manifold classification. This 

\begin{figure}[ht]
\centering
%\hspace{13pt}
  \begin{subfigure}{0.2\textwidth}
  \centering
  % include first image
  \includegraphics[width=\linewidth]{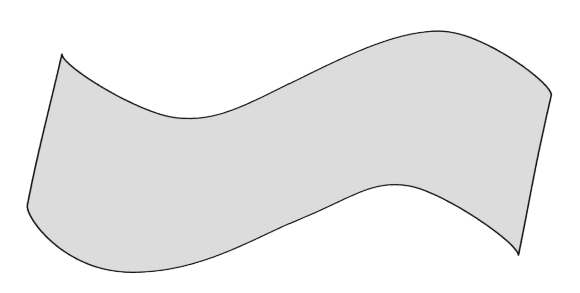}
  \caption{ }
  \label{fig:betti0}
  \end{subfigure}
  \begin{subfigure}{0.2\textwidth}
  \centering
  \includegraphics[width=\linewidth]{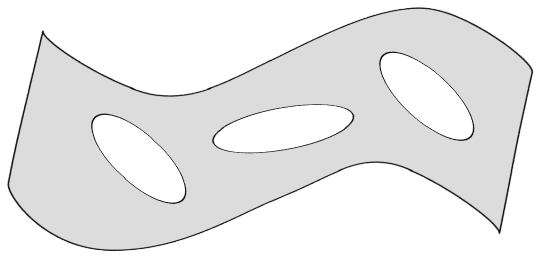}
  \caption{ }
  \label{fig:betti1}
  \end{subfigure}
  \begin{subfigure}{0.2\textwidth}
  \centering
  % include first image
  \includegraphics[width=\linewidth]{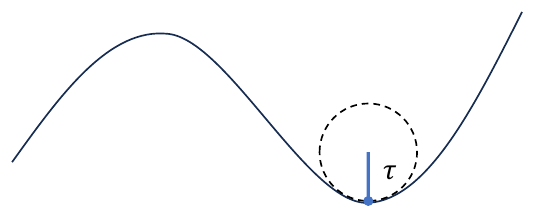}
  \caption{ }
  \label{fig:taul}
  \end{subfigure}
  \begin{subfigure}{0.2\textwidth}
  \centering
  \includegraphics[width=\linewidth]{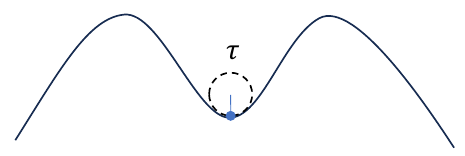}
  \caption{ }
  \label{fig:taus}
  \end{subfigure}
  \vspace{-0.1in}
  \caption{Illustration of \emph{Betti numbers} and \emph{reach}. (a) A 2-manifold embedding in $\R^3$ with $\beta_0=1,\beta_1=0$. (b) A 2-manifold embedded in $\R^3$ with $\beta_0=1,\beta_1=3$. (c) A 1-manifold with large reach. (d) A 1-manifold with small reach, which is the radius of the dashed circle. 
  %\cc{Maybe show case when $\beta_1=2$ or $3$? Also should show a case when $d=2$. When I read at first, I thought you only meant to address thickened 1-manifolds. Another question is what do you do with a figure-8? Is it still a manifold with $d=1$? If not, are you only dealing with 2 circles when $\beta_1=2$?}
  }
  \label{fig:betti}
  \vspace{-0.1in}
\end{figure}

Our theoretical result reveals for the first time how the topology, as a global structural characterization of data manifold, affects the network expressiveness. The beauty of the theorem is that it explicitly bounds the network size with both topology and geometry, giving us insights as to how the two important manifold properties affect learning.
To capture the manifold's topology, we use the classic \emph{Betti numbers}, which measure the number of connected components and holes within the manifold. 
For geometric measure, we use the \emph{reach} introduced by~\cite{federer1959curvature, amenta1998surface}, describing the manifold's overall flatness. 
%\cc{Explain condition number, and in the figure, you used reach instead of condition number.} 
See Figure \ref{fig:betti} for illustrations of these measures.
% Our theoretical analysis delivers an upper bound for network size controlled by both topology (the Betti numbers) and geometry (the condition number). 

%Specifically, our findings indicate that network size is quadratic with respect to the sum of the Betti numbers and polynomial in relation to the condition number.
Our main theoretical result is summarized informally below; the formal version is presented in Theorem 2.
\begin{main*}(Informal)
    Let $\gM\subset\R^D$ be a $d$-dimensional manifold ($d\le D$) from a family of thickened 1-manifold and has two classes. There exists a ReLU network classifier $g$ with depth at most $O(\log\beta+\log\frac{1}{\tau})$ and size at most $O(\beta^2+ (\frac{1}{\tau})^{d^2/2})$, such that with large probability, the true risk of $g$ is small. $\beta$ is the sum of Betti numbers and $\frac{1}{\tau}$ is the inverse of reach.
\end{main*}
According to our bound, the network size scales quadratically in terms of the sum of Betti numbers $\beta$.
%\cc{What is \textbf{total} betti number?}
%This can be validated by our empirical observations.
Conversely, in terms of \( \frac{1}{\tau} \), it scales as \( O\left((\frac{1}{\tau})^{d^2/2}\right) \). This bound reveals interesting insights. The growth of network size is only affected by Betti numbers quadratically. Meanwhile, the network size can be affected by condition number more significantly when the intrinsic manifold dimension is high. These insights can be the foundation for future development of tighter bounds, and potentially inspire new designs of network architectures that capitalize on data's intrinsic manifold structures.

In the following section, we discuss related works and compare our derived bounds with those previously established. In Section~\ref{sec:pre}, we define the problem and introduce the concept of the thickened $1$-manifold family. At first glance, this family may appear overly restrictive; however, we will explain why it actually represents a broad and versatile class of manifolds. In Section~\ref{sec:complexity}, we present our theoretical findings, detailing the step-by-step derivation of our bounds. Our results establish a new theoretical perspective that can stimulate further exploration into the expressiveness of networks. Looking ahead, this theory could inform the design of more efficient neural networks by leveraging insights from manifold topology and geometry.

    % \item To the best of our knowledge, this is the first work that includes both topology and geometry when determining the bounds of network size in manifold classification problem. \MY{tbc}
    % \item The bound we derive formally shows that within manifold classification, the geometric attributes of manifolds exert a dominant influence on the requisite network size. \MY{This is not proper as a contribution. Same reason as in abstract}
% This paper is structured in the following manner: Section~\ref{sec:related} reviews relevant literature, focusing on high-level comparisons between our approach and existing methodologies. Detailed comparisons, specifically regarding the mathematical bounds, are presented at the end of Section~\ref{sec:complexity}. Section~\ref{sec:assumption} provides a formal definition of our problem and outlines our manifold assumptions. In Section~\ref{sec:topocomplexity}, we elaborate on the latent manifold's formal definition and deduce the network size associated with topology. Section~\ref{sec:geocomplexity} calculates the overall network size for manifold classification, integrating the complexities from classifying the latent manifold and approximating the homeomorphism. Empirical validation of our approach is presented in Section~\ref{sec:experiment}.

\section{Related Works}
\label{sec:related}
%\myparagraph{Sample complexity.} While statistical learning can handle sample complexities, accommodating specialized manifold attributes like geometry and topology requires distinct methodologies. \cite{Niyogi2008Homology} provides a theoretic sample complexity bound to recover the homology of a sub-manifold. This bound is obtained in terms of a condition number that limits the curvature and nearness to self-intersection of the sub-manifold. \cite{samplecomplexNaray2010} provide bounds on sample complexity of Empirical Risk Minimization over a class of manifolds. Under this setting, they imply upper bounds on the sample complexity when testing the manifold hypothesis are independent of the ambient dimension, exponential in the intrinsic dimension, polynomial in the curvature and almost linear in the volume. Nevertheless, these works only focus on the sample complexity of manifolds, without considering how it might affect the hypothesis complexity, in our case the network size. Separately, \cite{narayanan2009sample} provides an insightful analysis into bounds on sample complexity required for classification using smooth decision boundaries, but it sidesteps any deep dive into the intricacies of decision boundary complexity.

\myparagraph{Network size with manifold geometry.} Multiple studies have formulated network size bounds across varied manifold learning contexts based on geometry. \cite{chartae2019stefan} establish a bound of $O(LdD\epsilon^{-d-d^2/2}(-\log^{1+d/2}\epsilon))$ on the network size for manifold reconstruction tasks. $L$ is the covering number in terms of the injectivity radius, a geometric property. They utilize an auto-encoder, denoted as \(D\circ E\), for the reconstruction of a manifold. Both the encoder \(E\) and the decoder \(D\) are designed to function as homeomorphisms. As a result, the overarching objective is the construction of a homeomorphism within the same space, which elucidates the absence of topological considerations in their outcomes. Our findings include not only the homeomorphism but also the classification network, with the latter being influenced by the manifold's topology.
\cite{lowmanifoldChen2019} demonstrate the existence of a network of size $O(\epsilon^{-d/n}\log\frac{1}{\epsilon}+D\log\frac{1}{\epsilon}+D\log D)$ that can approximate any smooth real function supported on a compact Riemannian manifold. In this context, $n$ denotes the order of smoothness of the function. Their primary objective is to illustrate that, in manifold learning, the manifold dimension chiefly determines network size, with only a marginal dependence on the ambient dimension. Moreover, their smoothness assumption is inapplicable to classification tasks, where the target function lacks continuity. Yet, the interplay between manifold properties and their impact on network size in manifold classification largely remains unexplored. 

\myparagraph{Classifier learned on manifold.} \cite{lsh2021dikkala} investigate network size in classification contexts. However, their foundational assumption is that a manifold's essence can be distilled into just two parameters: a centroid and a surrounding perturbation. They further assume there is a sensitive hashing property on manifolds. These assumptions are quite constrained, might not align with real-world complexities, and also overlooks the intrinsic properties of the manifold. Nevertheless, the aforementioned studies predominantly concentrate on network size and geometric traits, neglecting the equally critical role of topological features. \cite{buchanan2021deep} establish a lower bound on the size of classifiers for inputs situated on manifolds. However, their theoretical framework is restricted to smooth, regular simple curves; it fails to account for complex manifold structures. 
% This constraint negates the necessity to account for how variations in manifold topology might influence network size. 
\cite{guss2018characterizing} provide empirical evidence showing that classifiers, when trained on data with higher Betti numbers, tend to have slower convergence rates. They also highlight that with rising topological complexity, smaller networks face challenges in effective learning. These findings underscore the need for a more comprehensive theoretical understanding.
%Expanding upon the sample complexity findings, multiple studies have formulated network size bounds across varied manifold learning contexts. \MY{If it's okay to put the main theorem in the introduction, I will put the bound comparison here.} When it comes to approximating real functions set on manifolds, \cite{lowmanifoldChen2019} provide a bound showing that the network size expands exponentially in relation to the latent dimension, yet displays only a marginal dependence on the ambient dimension. However, their assumption for smooth functions cannot adapt to classification, where the desired function is not continuous. \cite{chartae2019stefan} established a network size outcome for manifold reconstruction using auto-encoders. Their approach involves segmenting the manifold into charts determined by the injectivity radius, and subsequently combining the capacities necessary to reconstruct each individual chart. While they claim their findings to capture both the geometric and topological facets of manifolds, their reliance on the condition number and injectivity radius — solely geometric features — does not show the relationship to topology. It is also noteworthy that their encoder and decoder are constructed as dual homeomorphisms, which is also why topology of manifold is not being considered. In contrast, the classification function is distinctly non-homeomorphic. 

There are some intriguing studies not primarily centered on manifold learning. Specifically, \cite{complexityneural2014bianchini} establish a bound for the Betti number of a neural network's expression field based on its capacity. Nevertheless, their proposed bound is loose, and it exclusively addresses the regions a network can generate, neglecting any consideration of input manifold. \cite{Safran2016DepthWidthTI} explore the challenge of approximating the indicator function of a unit ball using a ReLU network. While their primary objective is to demonstrate that enhancing depth is more effective than expanding width, their approach has provided valuable insights. \cite{naitzat2020topology} empirically examines the evolution of manifold topology as data traverses the layers of a proficiently trained neural network. We have adopted their concept of topological complexity. A number of studies, such as those by \cite{hanin2019complexity} and \cite{grigsby2022transversality}, concentrate on exploring the potential expressivity of neural networks. However, these works primarily focus on the network's inherent capabilities without extensively considering the characteristics of the input data.

%% file: preliminary.tex
\myparagraph{Topological Manifolds.} In this paper, we focus on a specific class of manifolds called the \emph{thickened 1-manifold family}, which are derived from operations on 1-manifolds. To lay the groundwork, we begin with the concept of a manifold. An n-dimensional manifold is a topological space where every point has a neighborhood that is homeomorphic (topologically equivalent) to $\R^n$. And an n-manifold with boundary allows for points that have neighborhoods homeomorphic to the half-space $\R_{\ge 0}\times\R^{n-1}$.  For example, a circle ($s^1$) is a 1-manifold without boundary, known as a closed manifold, while a line segment ($I^1$) is a 1-manifold with boundary. 

\begin{figure}[ht]
  \centering
  % include first image
  \includegraphics[width=\linewidth]{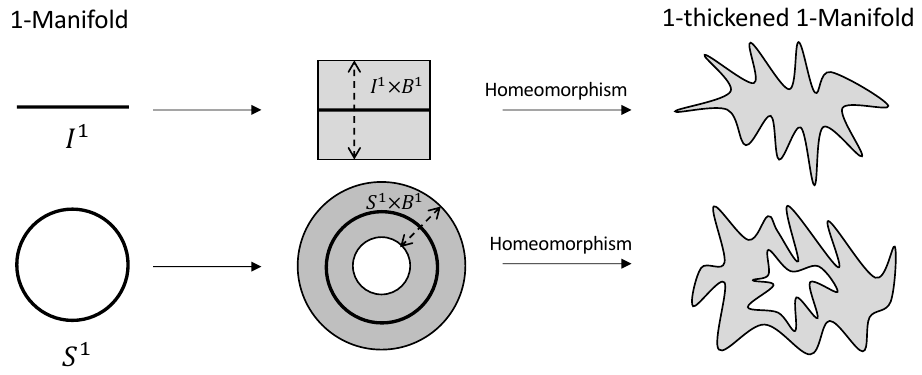}
  \caption{An illustration of the 1-thickened 1-manifold in 2D space. The top row shows a manifold that has the same homotopy type as a closed interval $I^1$, while the bottom row shows a manifold that is homotopy equivalent to a circle $S^1$. The 1-thickened 1-manifold family contains all manifolds that can be obtained through the disjoint union and connected sum of these manifolds.}
  \label{fig:1-manifold}
\end{figure}

\myparagraph{Thickened 1-manifold.} Our manifolds of interest are those obtained through operations on 1-manifolds, particularly when they are "thickened" to higher dimensions. Specifically, we consider \emph{$d$-thickened 1-manifolds}, which are $(d+1)$-dimensional manifolds homeomorphic to $\mathcal{M}^1 \times B^d$, where $\mathcal{M}^1$ is a compact 1-manifold (with or without boundary), and $B^d$ denotes a $d$-dimensional closed ball. \Figref{fig:1-manifold} illustrates the $1$-thickened $1$-manifold. According to the classification theorem for $1$-manifolds, $\gM^1$ is homeomorphic to either a closed interval $I^1$ or a circle $S^1$. Consequently, a 1-thickened 1-manifold is homeomorphic to either $I^1 \times B^1$ or $S^1 \times B^1$.

\myparagraph{Connected Sum and Disjoint Union.} To construct more complex manifolds from simpler ones, we utilize operations such as the connected sum and disjoint union. The connected sum of two closed manifolds $\gM$ and $\gN$, denoted $\gM \mathrel{\#} \gN$, is formed by removing a small open ball from each manifold and then gluing them together along the resulting boundary sphere. The disjoint union $\gM \sqcup \gN$ combines two manifolds by considering them together without connecting them. These operations yield new compact manifolds that combine the topological features of the original manifolds. Although the connected sum of thickened 1-manifolds differs from the standard definition of the connected sum for closed manifolds mentioned earlier, we utilize the fact that the boundary $\partial \mathcal{M}$ of a thickened 1-manifold $\mathcal{M}$ is a closed manifold. The connected sum of two thickened 1-manifolds is defined by performing the connected sum on their boundaries and extend the gluing to the interiors of identifying corresponding points in the neighborhoods adjacent to the removed sphere. The formal definition of connected sum is provided in \secref{sec:A-def}. \Figref{fig:sum} illustrates the connected sum between thickened 1-manifolds.

With the tools defined above, we can now define the thickened 1-manifold family, which is the class of manifolds we will study in this paper.

\begin{definition}[Thickened $1$-Manifold Family]
\label{def:1-manifolds}
Let $\gM^1$ represent a compact $1$-manifold (with or without boundary). A $d$-thickened $1$-manifold is a $(d+1)$-manifold homeomorphic to $\gM^1 \times B^d$. The family of $d$-thickened 1-manifolds, denoted by $\sM$, includes all such manifolds obtained via finite operations of connected sum and disjoint union.
\end{definition}

\myparagraph{Betti numbers.} We employ \textit{Betti number} $\beta_k(\gM)$ to quantify topology of a manifold $\gM$. $k$ is the dimension of that Betti number. 0-dimension Betti number $\beta_0(\gM)$ is the number of connected components in $\gM$, and $\beta_k(\gM)$ ($k\ge1$) can be informally described as the number of $k$-dimensional holes. $1$-dimensional hole is a circle and $2$-dimensional hole is a void. For the sake of coherence, we defer the formal definition of Betti numbers to Appendix~\ref{sec:A-def}. Following \cite{naitzat2020topology}, we utilize the total Betti number of $\gM$ as its topological complexity.
\begin{definition}[Topological Complexity]
    \label{def:topocomplexity}
    $\gM$ is a $d$-dimensional manifold. $\beta_k(\gM)$ is the $k$-dimensional Betti number of $\gM$. The topological complexity is defined as
    \begin{equation}
        \beta(\gM) = \sum_{k=0}^{d-1} \beta_k(\gM).
    \end{equation}
\end{definition}

\myparagraph{Reach and conditional number.} We then introduce metrics that encapsulate these geometric properties. For a compact manifold $\gM$, the reach $\tau$ is the largest radius that the open normal bundle about $\gM$ of radius $\tau$ is embedded in $\R^d$, i.e., no self-intersection. 
\begin{definition}[Reach and Condition Number]
    For a compact manifold $\gM\subset\R^D$, let
    \begin{equation}
    G = \{\rvx\in\R^D | \exists \rvp,\rvq\in\gM, \rvp\ne \rvq, ||\rvx-\rvp||=||\rvx-\rvq||=\inf_{\rvy\in\gM}||\rvx-\rvy||\}.
\end{equation}
The \textbf{reach} of $\gM$ is defined as $\tau(\gM)=\inf_{\rvx\in\gM, \rvy\in G}||\rvx-\rvy||$. The \textbf{condition number} $\frac{1}{\tau}$ is the inverse of the reach.
\end{definition}

\cite{Niyogi2008Homology} prove that the condition number controls the curvature of the manifold at every point. A modest condition number $1/\tau$ signifies a well-conditioned manifold exhibiting low curvature.

\myparagraph{Problem setup.} In this paper, we examine the topology and geometry of manifolds in the classification setting. We have access to a training dataset $\{(\rvx_i, y_i)|\rvx_i\in\gM, y_i\in[L]\}_{i=1}^n$, where $\gM=\bigsqcup_{l=1}^L\gM_l$. Each sample is drawn i.i.d. from a mixture distribution $\mu$ over $L$ disjoint manifolds with the corresponding label. For the simplification of notation, we build our theory on binary classification. It can be extended to multi-class without efforts in a one-verses-all setting. In binary case, the dataset is $\{(\rvx_i, y_i)|\rvx_i\in\gM, y_i\in\{0,1\}\}_{i=1}^n$, where $\gM=\gM_1\sqcup\gM_0\in\R^D$. $\gM_1$ and $\gM_0$ are two disjoint $d$-dimensional manifolds representing two classes. The label $y_i$ is determined by the indicator function
\begin{equation}
I_{\gM_1}(\rvx) =
\begin{cases}
1, &\rvx\in \gM_1, \\
0, &\text{otherwise}.
\end{cases}
\end{equation}
A neural network $h(\rvx):\R^D\to[0,1]$ approaches the classification problem by approximating the indicator function $I_{\gM_1}(\rvx)$. In the scope of this study, we focus on neural networks utilizing the ReLU (Rectified Linear Unit) activation function.
\begin{definition}[Adapted from \cite{arora2018understanding}]
    A ReLU multi-layer feed-forward network $h:\R^{w_0}\to\R^{w_{k+1}}$ with $k+1$ layers is defined as
    \begin{equation}
        h(\rvx) = h_{k+1}\circ h_{k}\circ\dots\circ h_1(\rvx),
    \end{equation}
    where $h_i:\R^{w_{i-1}}\to\R^{w_i}, h_i(\rvx)=\sigma(W_i\rvx+b_i)$ for $1\le i\le k$ are ReLU layers, and $h_{k+1}:\R^{w_k}\to\R^{w_{k+1}}, h_{k+1}(\rvx)=W_{k+1}\rvx+b_{k+1}$ is a linear layer. $\sigma$ is the ReLU activation function. The \textbf{depth} of the ReLU network is defined as $k+1$. The \textbf{width} of the ReLU network is $\max\{w_1,\dots,w_k\}$. The \textbf{size} of the ReLU network is $\sum_{i=1}^kw_i$.
\end{definition}

The approximation error of a ReLU network is determined by the true risk.
\begin{definition}[Approximation Error]
Let's consider the indicator function $I_{\gM_1}$ for a manifold $\gM_1$ in a binary classification problem where $\gM=\gM_1\sqcup\gM_0$. A neural network operates as a function $h(\rvx):\gM\to\R$. The approximation error of the neural network $h$ is then defined as:
\begin{equation}
    \text{True Risk: } R(h) = \int_{\gM}(h-I_{\gM_1})^2\mu(\mathbf{x})d\mathbf{x}.
\end{equation}
 $\mu$ is any continuous distribution over $\gM$.
\end{definition}

%% file: complexity.tex
In this section, we explore how the topology of manifolds influence network size in classification scenarios. Our results, derived methodically through construction, follow two steps.
First, we approximate a homeomorphism between the input manifold and a latent one; second, we carry out classification within this latent manifold. This latent manifold is designed to have simple geometric features, akin to those found in spheres and tori, while retaining the intrinsic topological characteristics of the original manifold. By design, the first phase is purely geometric, as the topological traits remain unaltered, while the subsequent classification phase is predominantly topological. Consequently, the required network size can be delineated into two distinct parts. We employ \textit{Betti numbers} and the \textit{condition number} as metrics to gauge topological and geometric complexities, respectively. Specifically, Betti numbers quantify the number of connected components and holes within the manifold, whereas the condition number characterizes the manifold's overall curvature. 

%In this section, we explore the complexities associated with using neural networks for manifold classification. Our aim is to gain a detailed understanding of how the manifold's topology and geometry influence the overall complexity of the network. Solid manifolds whose dimension is $1$ are intervals in the latent space (a real line). They possess only trivial topological properties as connected components (number of intervals). Thus, we focus on the non-trivial solid manifolds with 2-dimension or 3-dimension. To study the topological complexity, we must temporally discard the geometric properties of a manifold, such as curvature, diameter, and the like. For further elaboration, we focus on a specific subset of manifolds, which we term as `topological representatives', where the geometric properties appears trivial. The examination of geometric properties is carried out by constructing the homeomorphism between a manifold and its corresponding topological representative.

\subsection{Complexity Arising from Topology}
\label{sec:topocomplexity}
To focus on the topological aspects rather than the geometric intricacies of the manifold, our attention shifts to the elemental shapes that typify the thickened 1-manifold collection. Among these shapes are the $d$-dimensional balls, denoted as $B^d_r(\mathbf{c})$, and the solid $d$-tori, represented by $T^d_{r,R}$. The $d$-dimensional ball, $B^d_r(\mathbf{c})$, is characterized by a radius $r$ and is centered at $\mathbf{c}$, mathematically defined as
\begin{equation}
B^d_r(\mathbf{c})=\left\{\mathbf{x}\in\mathbb{R}^d : \Arrowvert\mathbf{x}-\mathbf{c} \Arrowvert_2 \le r\right\}.
\end{equation}
On the other hand, the solid $d$-torus, $T^d_{r,R}$, embodies a genus-1 torus with a tunnel radius of $r$ and a tunnel center radius of $R$, centered at $\mathbf{c}$. Its formula is given by
\begin{equation}
T^d_{r,R}(\rvc) = \left\{\mathbf{x}\in\mathbb{R}^d : (x_d-c_d)^2+\left(\sqrt{\sum_{i=1}^{d-1}(x_i-c_i)^2}-R\right)^2 \le r^2\right\}.
\end{equation}
It's pertinent to acknowledge that the general structure of a $d$-torus becomes significantly more complex as $d$ increases. The provided equation for $T^d_{r,R}$ represents only one of its potential configurations. Nevertheless, this depiction suffices within the context of the thickened 1-manifold ensemble.

\begin{lemma}[Topological Representative]
\label{lemma:toporepresentative}
    Let $\gM\subset\sM$ be a $d$-dimensional manifold from the thickened 1-manifold family. There exist a set of $m_1$ $d$-balls $\gB=\{B_{r_i}^d(\rvc_i)\}_{i=1}^{m_1}$ and a set of $m_2$ solid $d$-tori $\gT=\{T_{r_i,R_i}^d(\rvc_i)\}_{i=1}^{m_2}$, such that $\gM$ is homeomorphic to the union $(\bigcup_{B\in\gB}B)\cup(\bigcup_{T\in\gT}T)$, where $m_1+m_2\le\beta(\gM)$ is a constant integer. We term $\gM'=(\bigcup_{B\in\gB}B)\cup(\bigcup_{T\in\gT}T)$ as the \textbf{topological representative} of $\gM$. 
\end{lemma}
Given the classification theorem of 1-manifolds, compact 1-manifolds without boundary are homeomorphic to a circle, while compact 1-manifolds with boundary are homeomorphic to a closed interval. Therefore, the thickened 1-manifolds are homeomorphic to either a solid torus or a ball. Our preliminary analysis focuses on the network size associated with $d$-balls and $d$-tori. Using this as a foundation, we then explore how various topological configurations of thickened 1-manifold impact the size of neural networks. Proposition~\ref{prop:Rd-ball-est} determines the network size required to approximate a \(\R^d\) ball. While the original result is found in \cite{Safran2016DepthWidthTI}, our study utilizes fewer parameters and offers a different way to approximate the threshold function. Proposition~\ref{prop:3d-torus-est} outlines a network size bound for the approximation of a solid torus.
\begin{proposition}[Approximating a $\R^d$ Ball, adapted from Theorem 2 in \cite{Safran2016DepthWidthTI}]
\label{prop:Rd-ball-est}
Given $\epsilon>0$, there exists a ReLU network $h: \sR^d\to\sR$ with $3$ layers and with size at most $4d^2r^2/\epsilon + 2d + 2$, which can approximate the indicator function $I_{B^d_r}$ within error $R(h)\le\epsilon$ for any continuous distribution $\mu(\rvx)$. 
\end{proposition}

\begin{proposition}[Approximating a Solid Torus]
\label{prop:3d-torus-est}
Given $\epsilon>0$, there exists a ReLU network $h: \sR^d\to\sR$ with $5$ layers and with size at most $\frac{2d}{\epsilon}(4(d-1)(R+r)^2+8r^2+\frac{r}{\sqrt{R-r}})+9$, which can approximate the indicator function $I_{T_{r,R}^d}$ within error $R(h)\le\epsilon$ for any continuous distribution $\mu(\rvx)$.
\end{proposition}
Both proofs begin by expressing the indicator function as a composition of truncated power functions. Each truncated power function can be approximated by a piecewise linear function, which can be precisely represented by a ReLU layer. The composition of these power functions is then thresholded, and this threshold function can also be approximated by another ReLU layer. The overall approximation error is estimated across these two steps. Proposition \ref{prop:Rd-ball-est} and \ref{prop:3d-torus-est} address the network size associated with approximating $B_r^d$ and $T_{r,R}^d$. Detailed proofs can be found in the Appendix~\ref{sec:A-fundamental}. Building on this, we can infer the network size for approximating topological representatives by combining the complexities of approximating $B_r^d$ and $T_{r,R}^d$ with those of union operation. This consolidated insight is captured in Theorem~\ref{thm:topocomplex}.

\begin{theorem}[Complexity Arising from Topology]
\label{thm:topocomplex}
    Suppose $\gM'$ is the topological representative of $d$-manifold from the thickened $1$-manifold family. Given $\epsilon>0$, there exists a ReLU network $h:\R^d\to\R$ with depth at most $O(\log\beta)$ and size at most $O(\frac{d^2\beta^2}{\epsilon})$, that can approximate the indicator function $I_{\gM'}$ with error $R(h)\le\epsilon$ for any continuous distribution $\mu$ over $\R^d$. $\beta$ is the topological complexity of $\gM'$.
\end{theorem}
Since a topological representative is the union of balls and solid tori, the overall network size is computed by summing the network sizes required to approximate the balls and tori, along with the network size needed for the union operations. The detailed proof can be found in Appendix~\ref{sec:A-representative}. This theorem offers an upper bound on the network size required to approximate the indicator function of a topological representative. It is important to note that this captures the full range of complexities arising from the topology of a manifold $\gM\in\sM$, given that \(\gM\) and \(\gM'\) are homeomorphic. To the best of our knowledge, this is the first result bounding neural network size in terms of a manifold's Betti numbers.

\subsection{Overall Complexity}
\label{sec:geocomplexity}

For general manifolds in $\gM\in\sM$, due to their inherent complexity, often defy explicit expression. This hinders the direct use of function analysis for approximating their indicator functions, as was done in previous studies. To tackle this issue, we construct a homeomorphism, a continuous two-way transformation, between $\gM$ and its corresponding topological representative $\gM'$. This method only alters the geometric properties, preserving the object's topological attributes. Therefore, the network size in approximating the indicator function of $\gM$ is constructed by the approximation of the homeomorphism and the classification of topological representatives. The network size of constructing the homeomorphism is exclusively influenced by the geometric properties, whereas the network size of classifying topological representatives pertains solely to topological properties. The latter we already figured in previous section. This methodology enables us to distinguish the influence of topology and geometry of manifold on classifiers. In this section, we aim to obtain the overall network size for a classifier.

To build a homeomorphism from $\gM$, we first need to recover the homology of $\gM$. The subsequent proposition outlines a lower limit for the number of points essential to recover the homology of the initial manifold $\gM$.
\begin{proposition}[Theorem 3.1 in \cite{Niyogi2008Homology}]
\label{prop:samplecomplex}
    Let $\gM$ be a compact $d$-dimensional submanifold of $\R^D$ with condition number $1/\tau$. Let $X=\{\rvx_1,\rvx_2,..\rvx_n\}$ be a set of $n$ points drawn in i.i.d. fashion according to the uniform probability measure on $\gM$. Let $0<\epsilon<\frac{\tau}{2}$. Let $U=\bigcup_{\rvx\in X}B_{\epsilon}(\rvx)$ be a corresponding random open subset of $\R^D$. Then for all 
    \begin{equation}
    \label{eq:n}
        n>\lambda_1(log(\lambda_2)+log(\frac{1}{\delta})),
    \end{equation}
    $U$ is a $\epsilon$-cover of $\gM$, and the homology of $U$ equals the homology of $\gM$ with high confidence (probability $>1-\delta$).
    Here 
    \begin{equation}
    \label{eq:lambda}
        \lambda_1=\frac{vol(\gM)}{(cos^d\theta_1)vol(B_{\epsilon/4}^d)} \text{ and } \lambda_2 = \frac{vol(\gM)}{(cos^d\theta_2)vol(B_{\epsilon/8}^d)},
    \end{equation}
     $\theta_1=arcsin(\epsilon/8\tau)$ and $\theta_2=arcsin(\epsilon/16\tau)$. $vol(B_{\epsilon}^d)$ denotes the $d$-dimensional volume of the standard $d$-dimensional ball of radius $\epsilon$. $vol(\gM)$ is the $d$-dimensional volume of $\gM$.
\end{proposition}
This result stipulates a lower bound for the training set size necessary to recover the homology of the manifold, which is the foundation to learn the homeomorphism between a manifold $\gM\in\sM$ and its topological representative $\gM'$. However, directly constructing this homeomorphism remains challenging. As a workaround, we develop a simplicial homeomorphism to approximate the genuine homeomorphism. Notably, this simplicial approach lends itself readily to representation via neural networks.

\begin{figure}[ht]
    \centering
    \includegraphics[width=.9\linewidth]{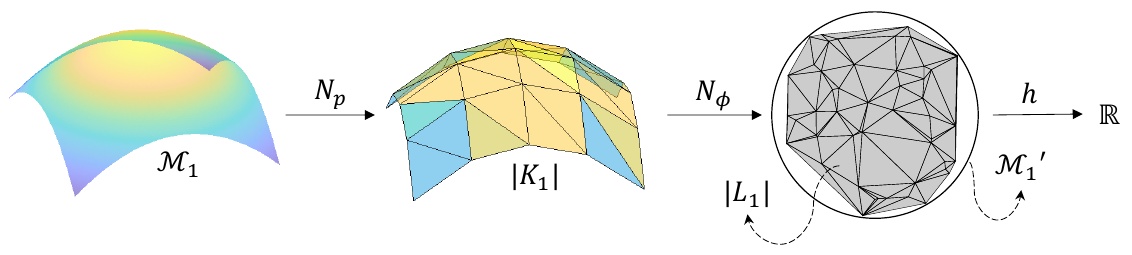}
    \caption{Construction of the network $g$. The network first learns a low-dimensional embedding and then performs classification in the embedding space. This paradigm mirrors the typical operation of deep networks. While the diagram illustrates only the process for the manifold of the positive class, the procedure for \( \gM_0 \) mirrors this operation identically.}
    \label{fig:construction}
\end{figure}

Combined with the topological representative classification network in Theorem~\ref{thm:topocomplex}, we can construct a classification network for general manifolds in $\sM$, as depicted in Figure~\ref{fig:construction}. Initially, we project \( \gM \) to its simplicial approximation \( |K| \) using a neural network \( N_p \). This is succeeded by a network \( N_{\phi} \) that facilitates the simplicial homeomorphism between \( |K| \) and \( |L| \), the latter being the simplicial approximation of the topological representative \( \gM' \). Finally, a network \( h \) is utilized to classify between \( |L_1| \) and \( |L_0| \). Consequently, the network's size is divided into two main parts: one focused on complexities related to geometric attributes and the other concerning topological aspects. This distinction separates topology from geometry in classification problems. 

In Theorem~\ref{thm:totalcomplexity}, we design such a neural network based on this training set, ensuring that approximation errors are effectively controlled. The detailed proof is provided in Appendix~\ref{sec:A-overallcomplex}. Our proof strategy begins with the construction of a ReLU network, followed by an evaluation of the network's size. Subsequently, we place bounds on the involved approximation errors.

\begin{theorem}[Main Theorem]
\label{thm:totalcomplexity}
    Let $\gM=\gM_1\sqcup\gM_0\subset\R^D$ be a $d$-dimensional manifold from the thickened $1$-manifold family. $\gM_1$ and $\gM_0$ are two disjoint sub-manifolds of $\gM$ representing two classes. The condition number of $\gM$ is $\frac{1}{\tau}$ and the total Betti number of $\gM_1$ is $\beta$. Given a training set $\{(\rvx_i, y_i)|\rvx_i\in\gM, y_i\in\{0,1\}\}_{i=1}^n$, where $\rvx_i$ are sampled i.i.d. from $\gM$ by a uniform distribution, and $y_i=I_{\gM_1}(\rvx_i)$. For any $\delta>0$, if inequality~(\ref{eq:n}) holds, then for any $\epsilon>0$, there exists a ReLU network $g$ with depth at most $O(\log\beta+d\log\frac{1}{\tau} + \log\log\frac{1}{\tau\delta})$ and size at most $O(\frac{d^2\beta^2}{\epsilon}+\tau^{-d^2/2}\log^{d/2}\frac{1}{\tau\delta}+D\tau^{-d}\log\frac{1}{\tau\delta})$, such that
    \begin{equation}
       P(R(g)\le\epsilon)>1-\delta,
    \end{equation}
    where $R(g)=\int_{\gM}(g-I_{\gM_1})^2\mu(\mathbf{x})d\mathbf{x}$ with any continuous distribution $\mu$.
\end{theorem}
\textit{Proof Sketch.} Since $\gM=\gM_1\sqcup\gM_0$ is from thickened $1$-manifold family, it has a topological representative $\gM'=\gM'_1\sqcup\gM'_0\subset \R^d$, where $\gM'_1$ and $\gM'_0$ are topological representatives of $\gM_1$ and $\gM_2$, respectively. The proof follows Figure~\ref{fig:construction}, by first constructing simplicial approximations $|K|$ and $|L|$ of $\gM$ and $\gM'$, respectively. Then we represent a simplicial homeomorphism $\phi:|K|\to |L|$ by a neural network $N_{\phi}$, where $K$ is constructed from $\gM$ and $L$ from $\gM'$. Built on the top of this, a projection from $\gM$ to its simiplicial approximation $|K|$ is represented by another network $N_p$. The overall network can be constructed by $g=h\circ N_{\phi}\circ N_p$. The proof is completed by first constructing the network $g$, and then bounding the approximation error.

Upon examining the depth and size of the neural network, it becomes evident that the topological complexity, denoted by \(\beta\), and the geometric complexity, symbolized by \(\tau\), are distinctly delineated. The topological complexity contributes \(O\left(\frac{d^2\beta^2}{\epsilon}\right)\) to the overall network size. In contrast, geometry contributes \(O\left(\tau^{-d^2/2}\log^{d/2}\frac{1}{\tau\delta} + D\tau^{-d}\log\frac{1}{\tau\delta}\right)\). %Notably, while the former scales quadratically with \(\beta\), the latter grows exponentially in relation to \(\frac{1}{\tau}\). This stark difference underscores the dominant influence of geometric complexity over its topological counterpart.

% It is important to note that our result is constructed as an upper bound. In practical scenarios, a neural network trained without specific constraints might not follow a strict sequence of first learning a homeomorphism to latent representations and then executing classification. Instead, it could adopt a more integrated approach, intertwining classification information during the representation learning process. This means that the actual network size could be significantly less than our provided bound. However, the tightness of topological bound \(O\left(\frac{d^2\beta^2}{\epsilon}\right)\) can be empirically verified with fixed dimension. We delve into this in the subsequent section.

%% file: discussion.tex
In this study, we delved into the intricate relationship between network size, and both geometric and topological characteristics of manifolds. Our findings underscored that while many existing studies have been focused on geometric intricacies, it is important to also appreciate the manifold's topological characteristics. These characteristics not only offer an alternative perspective on data structures but also influence network size in significant ways.

 Our proposed network size bounds represent theoretical upper limits, meaning that real-world implementations may yield efficiencies beyond these confines. To attain a more direct and refined theoretical bound, we may need more comprehensive descriptors of manifolds that go beyond merely the Betti numbers and the condition number. We leave this exploration for future work. We hope that our study acts as a catalyst for further research, pushing the boundaries of manifold learning and its applications in modern AI systems.

%% file: appendix.tex
\setcounter{proposition}{0}
\setcounter{theorem}{0}
\numberwithin{equation}{section}
\setcounter{equation}{0}
\numberwithin{figure}{section}
\setcounter{figure}{0}
\numberwithin{table}{section}
\setcounter{table}{0}

In this section, we formally prove the theoretical findings presented in the primary manuscript. Initially, we utilize some necessary definitions and existing results. Then we prove the network size bound for fundamental solid manifolds and general topological representatives.

\subsection{Additional Definitions}
\label{sec:A-def}
The first definition is \textit{Betti number}, which is a vital part of this paper. The $k$-th Betti number is defined as the rank of $k$-th homology group. Therefore, we have to properly define homology group first. Our definition follows \cite{hatcher2002algebraic} but is tailed for simplification. We first define \textit{simplicial homology} for simplicial complexes. (Actually for $\Delta$-complexes. we simplify it to avoid introducing $\Delta$-complexes.) Then extend it to \textit{singular homology} that can be applied to manifolds.

\myparagraph{Simplicial Homology.} Let $K$ be a simplicial complex, and let $K^k$ be the set of all $k$-dimensional simplices in $K$. The set of $K^k$ together with the field $\mathbb{Z}_2$ forms a group $C_k(K)$. It is a vector space defined on $\mathbb{Z}_2$ with $K^k$ as a basis. The element of $C_k(K)$ is called a $k$-chain. Let $\sigma\in K^k$ be a $k$-simplex. The boundary $\partial_k(\sigma)$ is the collection of its $(k-1)$-dimensional faces, which is a $k-1$-simplicial complex. The boudnary operator is linear, i.e.
\begin{equation*}
    \partial_k(z_1\sigma_1+z_2\sigma_2) = z_1\partial_k(\sigma_1) + z_2\partial_k(\sigma_2).
\end{equation*}
The boundary operator $\partial_k:C_k(K)\to C_{k-1}(K)$ introduces a chain complex
\begin{equation*}
    \cdots \xrightarrow[]{} C_{d} \xrightarrow[]{\partial_d} C_{d-1} \xrightarrow[]{\partial_{d-1}} C_{d-2} \xrightarrow[]{} \cdots \xrightarrow[]{} C_0 \xrightarrow[]{\partial_{0}} \emptyset.
\end{equation*}
$d$ is the maximum dimension of $K$. $\text{Ker}\,\partial_k$ is the collection of $k$-chains with empty boundary and $\text{Im}\,\partial_k$
is the collection of $(k-1)$-chains that are boundaries of $k$-chains. Then we can define the $k$-th homology group of the chain complex to be the quotient group $H_k=\text{Ker}\,\partial_k/\text{Im}\,\partial_{k+1}$. The $k$-th Betti number is defined by
\begin{equation*}
    \beta_k = \text{rank}\,H_k.
\end{equation*}

\myparagraph{Singular Homology.} Given a topological space \(X\), the \(k\)-th singular chain group \(C_k(X)\) is defined as the free Abelian group generated by the continuous maps \(\phi: K^k \to X\), where \(K^k\) is the standard \(k\)-simplex in \(\mathbb{R}^k\). Each such map is referred to as a singular \(k\)-simplex in \(X\).

A boundary operator \(\partial_k: C_k(X) \to C_{k-1}(X)\) can be defined as:
\[
\partial \phi = \sum_{i=0}^{n} (-1)^i \phi |_{[v_0,\cdots,\hat{v}_i,\cdots,v_n]},
\]
where \(\phi |_{[v_0,\cdots,\hat{v}_i,\cdots,v_n]}\) represents the restriction of \(\sigma\) to the \(i\)-th face of \(K^k\).

The \(k\)-th singular homology group \(H_k(X)\) is then represented as the quotient:
\[
H_k(X) = \text{Ker}\,\partial_k/\text{Im}\,\partial_{k+1}.
\]
The $k$-th Betti number is still defined as \(\beta_k = \text{rank}\,H_k(X)\).

\myparagraph{Connected Sum of Thickened 1-Manifold.} Let $\gM=\gM^1\times B^d$ and $\gN=\gN^1\times B^d$ be two $d$-thickened 1-manifolds, where $\gM^1$ and $\gN^1$ are compact 1-manifolds (with or without boundary), and $B^d$ is a $d$-dimensional closed ball. We define the \emph{connected sum} $\gM \# \gN$ by performing connected sum on their boundaries and then extend the gluing to the interior. The process is defined as:
\begin{enumerate}
    \item The connected sum of boundaries. The boundaries of $\gM$ and $\gN$ are closed manifolds, where $\partial\gM=\gM^1\times S^{d-1}$ and $\partial\gN=\gN^1\times S^{d-1}$. Two open neighborhoods $U \subset \partial\gM$ and $V\subset\partial\gN$ are removed from the boundaries. Then a homeomorphism $h:\partial U \to \partial V$ is defined to match the resulting boundaries. Then glue $\partial\gM \setminus U$ and $\partial\gN \setminus V$ along $\partial U$ and $\partial V$ via $h$, forming $\partial (\gM\#\gN)$.
    \item Extend the gluing to the interiors of $\gM$ and $\gN$ by identifying corresponding points in the neighborhoods adjacent to $U$ and $V$.
\end{enumerate}

\begin{figure}[ht]
  \centering
  \includegraphics[width=\linewidth]{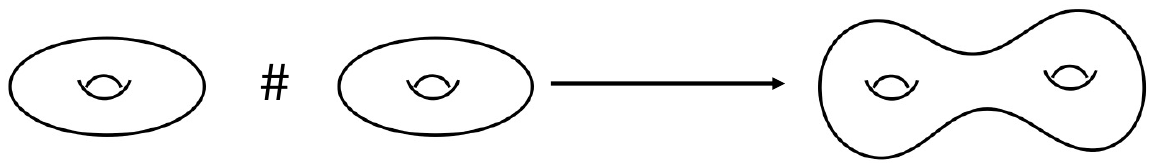}
  \caption{Illustration of connected sum.}
  \label{fig:sum}
\end{figure}

\subsection{Preliminary Results}
\label{sec:existing-lemmas}
We present some pre-established results regarding the network size associated with learning a 1-dimensional piecewise linear function, as well as basic combinations of functions.
\setcounter{lemma}{1}
\begin{lemma}[Theorem 2.2. in \cite{arora2018understanding}]
\label{lemma:pwl-est}
    Given any piecewise linear function $\sR\to\sR$ with $p$ pieces there exists a 2-layer ReLU network with at most $p$ nodes that can represent $f$. Moreover, if the rightmost or leftmost piece of a the piecewise linear function has $0$ slope, then we can compute such a $p$ piece function using a 2-layer ReLU network with size $p-1$.
\end{lemma}

\begin{lemma}[Function Composition, Lemma D.1. in \cite{arora2018understanding}]
\label{lemma:func-comp}
    If $f_1:\sR^d\to\sR^m$ is represented by a ReLU DNN with depth $k_1 + 1$ and size $s_1$, and $f_2 : \sR^m \to \sR^n$ is represented by a ReLU DNN with depth $k_2 +1$ and size $s_2$, then $f_2\circ f_1$ can be represented by a ReLU DNN with depth $k_1 +k_2 +1$ and size $s_1 + s_2$.
\end{lemma}

\begin{lemma}[Function Addition, Lemma D.2. in \cite{arora2018understanding}]
\label{lemma:func-add}
    If $f_1 : \sR^n \to \sR^m$ is represented by a ReLU DNN with depth $k+1$ and size $s_1$,and $f_2 :\sR^n \to \sR^m$ is represented by a ReLU DNN with depth $k+1$ and size $s_2$, then $f_1+f_2$ can be represented by a ReLU DNN with depth $k+1$ and size $s_1+s_2$.
\end{lemma}

\begin{lemma}[Taking maximums, Lemma~D.3. in \cite{arora2018understanding}]
\label{lemma:func-max}
    Let $f_1,...,f_m:\R^n\to\R$ be the functions that each can be represented by ReLU networks with depth $k_i+1$ and size $s_i$, $i=1,...,m$. Then the function $f:\R^n\to\R$ defined as $f=\max\{f_1,...,f_m\}$ can be represented by a ReLU network of depth at most $\max\{k1,...,k_m\}+\log(m)+1$ and size at most $s_1+...+s_m+4(2m-1)$.
\end{lemma}

We proceed to disclose the network size involved in learning a 1-dimensional Lipschitz function.
\begin{lemma}[Lipschitz Function Approximation, adapted from Lemma 11 in \cite{eldan2016power}]
\label{lemma:lipschitz-est}
For any L-Lipschitz function $f:\sR\to\sR$ which is constant outside a bounded interval $[a,b]$, and for any $\epsilon>0$, there exits a two-layer ReLU network $h(x)$ with at most $\lceil L(b-a)/\epsilon\rceil+1$ nodes, such that
\begin{equation*}
    \sup_{x\in\sR}|f(x)-h(x)|<\epsilon.
\end{equation*}
\end{lemma}
\begin{proof}
    We follow the original proving idea but adapt it for better understanding. We prove the lemma by estimate the Lipschitz function by a piece-wise linear function within error $\epsilon$ and use a two-layer ReLU network to represent the piece-wise linear function.

    We first cut the interval equally into $m$ sections $[a,b]=\bigcup_{i=1}^m[a+(i-1)\delta, a+i\delta]$, where $\delta=(b-a)/m$. For each interval $I_i = [a+(i-1)\delta, a+i\delta]$, we denote $f_i(x)=f|_{I_i}$. Then $\forall x_1, x_2 \in I_i$, $|f_i(x_1)-f_i(x_2)|\le L|x_1-x_2| \le L\delta$. Let $h_i(x)$ be the linear function defined on this interval and connect $(a+(i-1)\delta, f_i(a+(i-1)\delta))$ and $(a+i\delta, f_i(a+i\delta))$. Then we can bound the difference between $f_i(x)$ and $h_i(x)$ by
    \begin{align}
    \begin{split}
        |f_i(x)-h_i(x)| &\le \max\{|\max f_i(x) - \min h_i(x)|, |\min f_i(x) - \max h_i(x)|\} \\
        &= \max\{|\max f_i(x) - f_i(a+(i-1)\delta)|, |\min f_i(x) - f_i(a+i\delta))|\} \\
        &\le L\delta.
    \end{split}
    \end{align}
    The second line assumes $h_i(x)$ is non-decreasing. The other case can also be easily verified. By setting $m=\lceil \frac{L(b-a)}{\epsilon}\rceil$, for every interval, the error is controlled by $\epsilon$. Let $h(x)$ be the collection of all $h_i$ and also the constant outside of $[a,b]$, so we have $\sup_{x\in\sR} |f(x)-h(x)|<\epsilon$.

    $h(x)$ is a piece-wise linear function with $m+2$ pieces. According to Lemma~\ref{lemma:pwl-est}, there exists a 2-layer ReLU network with at most $m+1$ pieces that can represent $h(x)$. Proof done.
\end{proof}

\subsection{Approximating Fundamental Solid Manifolds}
\label{sec:A-fundamental}
Now we are in a good position to prove Proposition \ref{prop:Rd-ball-est} and \ref{prop:3d-torus-est}.

\begin{proposition}[Approximating a $\R^d$ Ball, adapted from Theorem 2 in \cite{Safran2016DepthWidthTI}]
Given $\epsilon>0$, there exists a ReLU network $h: \sR^d\to\sR$ with $3$ layers and with size at most $4d^2r^2/\epsilon + 2d + 2$, which can approximate the indicator function $I_{B^d_r}$ within error $R(h)\le\epsilon$ for any continuous distribution $\mu(\rvx)$. 
\end{proposition}
\begin{proof}
    We generally follow the original proof but derive a slightly different bound with fewer parameters. The proof is organized by first using a non-linear layer to approximate a truncated square function and then using another non-linear layer to approximate a threshold function.
    Consider the truncated square function
    \begin{equation}
        l(x;r) = \min\{x^2, r^2\}.
    \end{equation}
    Clearly $l(x;r)$ is a Lipschitz function with Lipschitz constant $2r$. Applying Lemma~\ref{lemma:lipschitz-est}, we have a 2-layer ReLU network $h_{11}$ that can approximate $l(x;r)$ with 
    \begin{equation}
        \sup_{x\in\R} \big|h_{11}(x)-l(x)\big| \le \epsilon_1,
    \end{equation}
    with at most $2r^2/\epsilon_1+2$ nodes. Now for $\rvx\in\R^d$, let
    \begin{equation}
    \label{eq:sum-dimension}
        h_1(\rvx) = \sum_{i=1}^d h_{1i}(x_i).
    \end{equation}
    Note that $h_1$ is also a 2-layer network because no extra non-linear operation is introduced in \eqref{eq:sum-dimension}, and has size at most $2dr^2/\epsilon_1+2d$. This can also be verified by Lemma~\ref{lemma:func-add}. Let 
    \begin{equation}
        L(\rvx) = \sum_i^d L(x_i; r),
    \end{equation}
    and we have
    \begin{equation}
        \sup_{\rvx} \big|h_1(\rvx) - L(\rvx) \big| \le d\epsilon_1.
    \end{equation}
    Let $\epsilon_1 = d\epsilon_1$, then $h_1$ has size at most $2d^2r^2/\epsilon_1+2d$. Although $L(\rvx)$ is different from $\sum x_i^2$, the trick here is to show $B_r^d=\{\rvx:L(\rvx)\le r^2\}$.

    On the one hand, if $L(rvx)\le r^2$, remember that
    \begin{equation}
        L(\rvx) = \sum_{i=1}^d \min\{x_1^2, r^2\}\le r^2.
    \end{equation}
    This means for all $x_i$, $x_i\le r^2$. Therefore, $L(\rvx)=\sum_{i=1}^d x_i^2$.
    On the other, $L(\rvx) > r^2$ only happens when there exists a $i$, such that $x_i^2>r^2$. Thus, $\rvx\notin B_r^d$. Consequently, one can represent $I_{B_r^d}$ by $L(\rvx)\le r^2$.

    The next step towards this proposition is to construct another 2-layer ReLU network to threshold $L(\rvx)$. Consider
    \begin{equation}
    \label{eq:threshold-est-ball}
      f(x) =
      \begin{cases}
        1, & x < r^2-\delta, \\
        \frac{r^2-x}{\delta}, &x\in [r^2-\delta, r^2],\\
        0, &  x > r^2.
      \end{cases}
    \end{equation}
    Note that $f$ is a 3-piece piece-wise linear function that approximates a threshold function. According to Lemma~\ref{lemma:pwl-est}, a 2-layer ReLU network $h_2$ with size 2 can represent $f$. The function $f\circ L(\rvx)$ can then be estimated by a 3-layer network $h=h_2\circ h_1$, whose size is $2d^2r^2/\epsilon_1+2d+2$. The next step is to bound the error between $h$ and $I_{B_r^d}$. We consider the $L_2$-type bound $||h(\rvx)-I_{B_r^d}(\rvx)||_{L_2(\mu)}=\int_{\R^d}(h(\rvx) - I_{B_r^d}(\rvx))^2\mu(\rvx)d\rvx$. We divide the integral into two parts
    \begin{align}
    \begin{split}
        &||h(\rvx)-I_{B_r^d}(\rvx)||_{L_2(\mu)} \\
        &\le ||f\circ L(\rvx) - I_{B_r^d}(\rvx)||_{L_2(\mu)} + ||f\circ L(\rvx) - h_2\circ h_1(\rvx)||_{L_2(\mu)} \\
        &= I_1 + I_2.
    \end{split}
    \end{align}

    Since $\mu(\rvx)$ is continuous, there exists $\delta$ such that
\begin{equation}
    \int_{S_{\delta}}\mu(\rvx)d\rvx \le \epsilon_2.
\end{equation}
$S_{\delta} = \{\rvx\in\sR^d: r^2-\delta \le \sum_{i=1}^d x_i^2 \le r^2\}$. Combine \eqref{eq:threshold-est-ball} we have
\begin{align}
\label{eq:I1}
    \begin{split}
    I_1 &= \int_{\sR^d} (f\circ L(\rvx) - I_{B_r^d}(\rvx))^2\mu(\rvx)d\rvx  \\
    &= \int_{S_{\delta}} (f\circ L(\rvx) - I_{B_r^d}(\rvx))^2\mu(\rvx)d\rvx \\
    & = \int_{S_{\delta}} (f\circ L(\rvx) - 1)^2\mu(\rvx)d\rvx \\
    & \le \int_{S_{\delta}}\mu(\rvx)d\rvx \\
    & \le \epsilon_2.
    \end{split}
\end{align}
The first inequality is because $f\in [0, 1]$, such that $(f\circ L(\rvx)-1)^2\le 1$.
The second part of the error can be easily bounded by its infinity norm.
\begin{equation}
\label{eq:I2}
    I_2 = ||f\circ L(\rvx) - h_2\circ h_1(\rvx)||_{L_2(\mu)} \le ||f\circ L(\rvx) - h_2\circ h_1(\rvx)||_{\infty} \le \epsilon_1.
\end{equation}
The last inequality is because $h_2$ is the exact representation of $f$, the error only occurs between $L(\rvx)$ and $h_1$.
Combine \ref{eq:I1} and \ref{eq:I2}, and let $\epsilon_1 = \epsilon_2 = \epsilon/2$, we have
\begin{equation}
    ||h(\rvx)-I(\rvx)||_{L_2(\mu)} \le \epsilon.
\end{equation}
The size of network $h$ is then bounded by $4d^2r^2/\epsilon + 2d + 2$.
\end{proof}

\begin{proposition}[Approximating a Solid Torus]
Given $\epsilon>0$, there exists a ReLU network $h: \sR^d\to\sR$ with $5$ layers and with size at most $\frac{2d}{\epsilon}(4(d-1)(R+r)^2+8r^2+\frac{r}{\sqrt{R-r}})+9$, which can approximate the indicator function $I_{T_{r,R}^d}$ within error $R(h)\le\epsilon$ for any continuous distribution $\mu(\rvx)$.
\end{proposition}
\begin{proof}
The proof is done by two steps. We first use layers to estimate a truncated function. Then estimate a threshold function by another layer.

Consider the truncated square function and root function,
\begin{align*}
    &l_1(x;\gamma) = \min\{x^2, \gamma^2\}, \\
    &l_2(x;\gamma_1,\gamma_2) = \min\{\max\{\sqrt{x}, \gamma_1\}, \gamma_2\}, (\gamma_1<\gamma_2).
\end{align*}
The Lipschitz constants for $l_1$ and $l_2$ are $2\gamma$ and $\frac{1}{2\sqrt{\gamma_1}}$, respectively. By Lemma~\ref{lemma:lipschitz-est}, there is a 2-layer ReLU network to approximate $l_1$ and $l_2$ with size $\lceil 4\gamma^2/\epsilon_1\rceil+1$ and $\lceil (\gamma_2-\gamma_1)/(2\epsilon_1\sqrt{\gamma_1})\rceil+1$, respectively. Let
\begin{equation}
    L(\rvx)=l_1(x_d;r)+l_1(l_2(\sum_{i=1}^{d-1} l_1(x_i;R+r);R-r, R+r) - R;r).
\end{equation}
Then it is time to show $T^d_{r,R} = \left\{\mathbf{x}\in\mathbb{R}^d : x_d^2+\left(\sqrt{\sum_{i=1}^{d-1}x_i^2}-R\right)^2 \le r^2\right\}=\{\rvx: L(\rvx)\le r^2\}$. 
For $\rvx\in I_{T^d_{r,R}}(\rvx)$, the following inequalities hold 
\begin{align}
\label{eq:inner}
    &x_i^2 \le (R+r)^2, 1\le i\le d-1, \\
\label{eq:middle}
    &R+r \ge\sqrt{\sum_{i=1}^{d-1}x_i^2}\ge R-r, \\
\label{eq:outter}
    &x_d^2 \le r^2, (\sqrt{\sum_{i=1}^{d-1}x_i^2}-R)^2 \le r^2.
\end{align}
These indicate that $L(\rvx)=x_d^2+\left(\sqrt{\sum_{i=1}^{d-1}x_i^2}-R\right)^2\le r^2$, when $\rvx\in T^d_{r,R}$. And when $\rvx\notin T^d_{r,R}$, if $L(\rvx)=x_d^2+\left(\sqrt{\sum_{i=1}^{d-1}x_i^2}-R\right)^2$ still holds, clearly $L(\rvx)>r^2$. Otherwise, one of the inequalities in \ref{eq:inner}, \ref{eq:middle} and \ref{eq:outter} must break. If one of \ref{eq:outter} breaks, then clearly $L(\rvx)> r^2$. If \ref{eq:middle} does not hold, then $(\sqrt{x_1^2+x_2^2}-R)^2>r^2$, resulting $L(\rvx)>r^2$. The violation of \ref{eq:inner} resulting violation of \ref{eq:middle}, which then leads to $L(\rvx)>r^2$.

To see how a ReLU network can estimate $L(\rvx)$, we start by estimating each of its component. We define the following 2-layer networks.
To make the overall network take $\rvx\in\sR^d$ as input, we consider the structure in figure~\ref{fig:netconstruction}.

\begin{figure}[tbh!]
\centering
\tikzstyle{line} = [draw, -latex']
\begin{tikzpicture}[thick, main/.style = {draw, circle}, minimum size=9mm]

\node[main] (1) {$x_1$}; 
\node[main] (12)[below=0.5cm of 1]{...};
\node[main] (13)[right=1.5cm of 12] {...};
\node[main] (2) [below=0.5cm of 12] {\tiny{$x_{d-1}$}};
\node[main] (3) [below=0.5cm of 2] {$x_d$};
\node[main] (4) [right=1.5cm of 1] {$h_{11}$};
\node[main] (5) [right=1.5cm of 2] {\tiny{$h_{1,d-1}$}};
\node[main] (6) [right=1.5cm of 3] {$x_d$};
\node[main] (8) [right=1.5cm of 6] {$x_d$};
\node[main] (7) [right=1.5cm of 13] {$h_2$};
\node[main] (9) [right=1.5cm of 7] {$h_{31}$};
\node[main] (10) [right=1.5cm of 8] {$h_{32}$};
\node[main] (11) [below right=0.5cm and 1.5cm of 9] {$+$};
\path[line] (12) -- (13);
\path[line] (13) -- (7);
\path[line] (1) -- (4);
\path[line] (2) -- (5);
\path[line] (3) -- (6);
\path[line] (4) -- (7);
\path[line] (5) -- (7);
\path[line] (6) -- (8);
\path[line] (7) -- node [text width=1cm,midway,above=0.5mm,align=center] {$-R$} (9);
\path[line] (8) -- (10);
\path[line] (9) -- (11);
\path[line] (10) -- (11);
\end{tikzpicture} 
\caption{Network construction.}
\label{fig:netconstruction}
\end{figure}

The size of parts in the network is provided in table~\ref{tab:netsize}.
\begin{table}[tbh!]
    \centering
    \caption{Sub-network Size.}
    \label{tab:netsize}
    \begin{tabular}{c|c|c}
       Network  & Target & Size \\
       \hline
        $h_{1i}$ & $l_1(x_i;R+r)$ & $s_{1i}=\lceil 4(R+r)^2/\epsilon_1\rceil+1$ \\
        $h_{2}$ & $l_2(x;R-r,R+r)$ & $s_2=\lceil r/(\epsilon_1\sqrt{R-r})\rceil+1$ \\
        $h_{31}$ & $l_1(x;,r)$ & $s_{31}=\lceil 4r^2/\epsilon_1\rceil+1$ \\
        $h_{32}$ & $l_1(x_d;,r)$ & $s_{32}=\lceil 4r^2/\epsilon_1\rceil+1$
    \end{tabular}
    %\caption{Caption}
    \label{tab:strcut}
\end{table}

By Lemma~\ref{lemma:func-add} and Lemma~\ref{lemma:func-comp} and the given structure, a ReLU network $\tilde{L}$ with depth $4$ and size $(d-1)s_{11}+s_2+s_{31}+s_{32}+2$, where $d=3$, can approximate $L(\rvx)$ such that
\begin{equation}
    \sup_{\rvx}|L(\rvx)-\tilde{L}(\rvx)|\le d\epsilon_1.
\end{equation}

The next step is to threshold $L(\rvx)$. Consider a function
\begin{equation}
\label{eq:threshold-est}
  f(x) =
  \begin{cases}
    1, & x < r^2-\delta, \\
    \frac{r^2-x}{\delta}, &x\in [r^2-\delta, r^2],\\
    0, &  x > r^2.
  \end{cases}
\end{equation}
This function approximates a thresholding function $I[x\le r^2]$ but with error inside the interval $[r^2-\delta, r^2]$. By Lemma~\ref{lemma:pwl-est}, a 2-layer ReLU network $\tilde{f}$ with size $2$ can represent $f(x)$. Then $\tilde{f}\circ\tilde{L}$ is a ReLU network with depth 5 and size $(d-1)s_{11}+s_2+s_{31}+s_{32}+4$, such that
\begin{equation}
    \sup_{\rvx}|f\circ L(\rvx)-\tilde{f}\circ\tilde{L}(\rvx)|\le\epsilon_1,
\end{equation}
with letting $\epsilon_1=\epsilon_1/d$.

Let $h(\rvx)=\tilde{f}\circ\tilde{L}(\rvx)$. We claim that $h(\rvx)$ is the desired network with depth 5 and size 
\begin{align}
\begin{split}
    &(d-1)s_{11}+s_2+s_{31}+s_{32}+4 \\
    &= \frac{d}{\epsilon_1}(4(d-1)(R+r)^2+8r^2+\frac{r}{\sqrt{R-r}})+9 \\
    &= O(\frac{d^2}{\epsilon_1}).
\end{split}
\end{align}
To finalize our proof, we just need to bound the error $||h(\rvx)-I_{T^d_{r,R}}(\rvx)||_{L_2(\mu)}$. The proof follows proof of Proposition~\ref{prop:Rd-ball-est}. The error is divided into two parts and is bounded separately. The only difference is we define $S_{\delta}$ to be
$S_{\delta} = \left\{\rvx\in\sR^d: r^2-\delta \le x_d^2+\left(\sqrt{\sum_{i=1}^{d-1}x_i^2}-R\right)^2 \le r^2\right\}$, such that
\begin{equation}
    \int_{S_{\delta}}\mu(\rvx)d\rvx \le \epsilon_2.
\end{equation}
We can get $I_1\le\epsilon_2$, and $I_2\le\epsilon_1$. Let $\epsilon_1 = \epsilon_2 = \epsilon/2$, we have
\begin{equation}
    ||h(\rvx)-I(\rvx)||_{L_2(\mu)} \le \epsilon.
\end{equation}
And $h$ has size at most $\frac{2d}{\epsilon}(4(d-1)(R+r)^2+8r^2+\frac{r}{\sqrt{R-r}})+9$.
\end{proof}

\subsection{Approximating Topological Representatives}
\label{sec:A-representative}
After getting the size arising from fundamental manifolds, we proceed to study the combination of them. We start by proving the representative property.
\setcounter{lemma}{0}
\begin{lemma}[Topological Representative]
    Let $\gM\subset\sM$ be a $d$-dimensional manifold from the thickened 1-manifold family. There exist a set of $m_1$ $d$-balls $\gB=\{B_{r_i}^d(\rvc_i)\}_{i=1}^{m_1}$ and a set of $m_2$ solid $d$-tori $\gT=\{T_{r_i,R_i}^d(\rvc_i)\}_{i=1}^{m_2}$, such that $\gM$ is homeomorphic to the union $(\bigcup_{B\in\gB}B)\cup(\bigcup_{T\in\gT}T)$, where $m_1+m_2\le\beta(\gM)$ is a constant integer. We term $\gM'=(\bigcup_{B\in\gB}B)\cup(\bigcup_{T\in\gT}T)$ as the \textbf{topological representative} of $\gM$.
\end{lemma}
\begin{proof}
Since $\gM \subset \sM$, regarding the definition, $\gM$ is homeomorphic to connect sum or disjoint union of $m$ thickened 1-manifolds. A thickened $1$-manifold is denoted as $\gM^1\times B^{d-1}$. And based on the classification theorem of $1$-manifold, $\gM^1$ is homeomorphic either to a circle $S^1$ or $I=[0,1]$. If $\gM^1\sim S^1$, then $\gM^1\times B^{d-1}\sim S^1\times B^{d-1}\sim T_{r,R}^d$. If $\gM^1\sim I$, then $\gM^1\times B^{d-1}\sim I\times B^{d-1}\sim B_r^d$. Now let $\gM_1$ and $\gM_2$ be two thickened $1$-manifold, the connected sum of $\gM_1$ and $\gM_2$, denoted by $\gM_1\oplus\gM_2$, can be represented by the union of two manifolds $\gM_1\cup\gM_2$. Since $\gM$ is homeomorphic to connect sum or disjoint union of $m$ thickened 1-manifolds, suppose $m_1$ of them are homeomorphic to $B_{r}^d$ and $m_2$ of them are homeomorphic to $T_{r,R}^d$. Therefore, $\gM\sim (\bigcup_{B\in\gB}B)\cup(\bigcup_{T\in\gT}T)$. 

Notice that among $\gB$, the union is disjoint union, otherwise the union is still homeomorphic to a $B_r^d$. Similarly, the union between $\gB$ and $\gT$ is also disjoint union, otherwise the union is still homeomorphic to a solid torus $T_{r,R}^d$. The union among $\gT$ could be joint union or disjoint union. For two solid tori $T_1^d$ and $T_2^d$, $\beta(T_1^d)=\beta(T_1^d)=2$. $\beta(T_1^d\cup T_2^d)\ge 3 > 2$. Overall, $\beta(\gM)\ge m_1+m_2$.
\end{proof}
\setcounter{lemma}{6}
\begin{lemma}[Manifold Union]
\label{lemma:combination}
    $\gM_1$ and $\gM_2$ are two manifolds in $\R^d$. $I_{\gM_1}$ can be approximated by a ReLU network $h_1$ with depth $d_1+1$ and size at most $s_1$ with error $R(h_1)<\epsilon_1$, $I_{\gM_2}$ can be approximated by a ReLU network $h_2$ with depth $d_2+1$ and size at most $s_2$ with error $R(h_2)<\epsilon_2$. Then $I_{\gM_1\cup\gM_2}$ can be approximated within error $\epsilon_1+\epsilon_2$ by a ReLU network with depth at most $\max\{d_1,d_2\}+2$ and size at most $s_1+s_2+2$.
\end{lemma}
\begin{proof}
    We represent $I_{\gM_1\cup\gM_2}=I_{x>0}\circ(I_{\gM_1}+I_{\gM_2})$ by a threshold function $I_{x>0}$. The threshold function can be approximate by a function
    \begin{equation}
        f(x) = 
        \begin{cases}
            0, &x\le 0 \\
            \frac{x}{\delta}, & x\in(0,\delta) \\
            1, &x\ge\delta.
        \end{cases}
    \end{equation}
    with errors only in $(0,\delta)$. $f$ can be represented by a 2-layer ReLU network $h_f$ with size $2$. Then if let $h=h_f\circ(h_1+h_2)$, according to Lemma~\ref{lemma:func-add} and \ref{lemma:func-comp}, $h$ is a neural network with depth $\max\{d_1,d_2\}+2$ and size $s_1+s_2+2$. Then we bound the error
    \begin{align}
    \begin{split}
        ||h-I_{\gM_1\cup\gM_2}||_{L_2(\mu)} &= ||h_f\circ(h_1+h_2)-I_{x>0}\circ(I_{\gM_1}+I_{\gM_2})||_{L_2(\mu)} \\
        &\le ||h_f\circ(h_1+h_2)-f\circ(I_{\gM_1}+I_{\gM_2})||_{L_2(\mu)} \\
        &+ ||f\circ(I_{\gM_1}+I_{\gM_2})-I_{x>0}\circ(I_{\gM_1}+I_{\gM_2})||_{L_2(\mu)} \\
        &\le ||h_f\circ(h_1+h_2)-h_f\circ(I_{\gM_1}+I_{\gM_2})||_{L_2(\mu)} \\
        &\le ||h_f||_{\infty}||(h_1+h_2)-(I_{\gM_1}+I_{\gM_2})||_{L_2(\mu)} \\
        &\le ||(h_1+h_2)-(I_{\gM_1}+I_{\gM_2})||_{L_2(\mu)} \\
        &\le \epsilon_1+\epsilon_2
    \end{split}
    \end{align}

\end{proof}

\begin{theorem}[Complexity Arising from Topology]
  Suppose $\gM'$ is the topological representative of $d$-manifold from the thickened $1$-manifold family. Given $\epsilon>0$, there exists a ReLU network $h:\R^d\to\R$ with depth at most $O(\log\beta)$ and size at most $O(\frac{d^2\beta^2}{\epsilon})$, that can approximate the indicator function $I_{\gM'}$ with error $R(h)\le\epsilon$ for any continuous distribution $\mu$ over $\R^d$. $\beta$ is the topological complexity of $\gM'$.
\end{theorem}
\begin{proof}
    Since $\gM'$ is a topological representative, according to Lemma~\ref{lemma:toporepresentative}, there exist a set of $m_1$ $d$-balls $\gB=\{B_{r_i}^d(\rvc_i)\}_{i=1}^{m_1}$ and a set of $m_2$ solid $d$-tori $\gT=\{T_{r_i,R_i}^d(\rvc_i)\}_{i=1}^{m_2}$, such that $\gM'=(\bigcup_{B\in\gB}B)\cup(\bigcup_{T\in\gT}T)$. Let $m=m_1+m_2$. According to proposition~\ref{lemma:combination}, $I_{\gM'}$ can be approximated by a ReLU network $h$ with depth at most $\max\{d_1,d_2,...,d_{m}\}+\log m$ and size at most $\sum_{i=1}^{m}s_i + \log m$, with error $R(h)\le\sum_{i=1}^{m}\epsilon_i$. Then according to Proposition~\ref{prop:Rd-ball-est} and \ref{prop:3d-torus-est}, $s_i\sim O(d^2/\epsilon_i)$, $d_i\sim O(1)$ and take $\epsilon_i$ to be all the same for all $i=[m]$. Let $\epsilon = m\epsilon_i$ and note that $m\le\beta$. We have $h$ has depth at most $O(\log \beta)$ and size at most $O(\frac{d^2\beta^2}{\epsilon})$, and can approximate $I_{\gM'}$ with error $R(h)\le\epsilon$.
\end{proof}

\subsection{Overall Complexity}
\label{sec:A-overallcomplex}
We present a result from \cite{GonzalezDiaz2019TwohiddenlayerFN}, which gives a bound of network size to represent a simplicial map.

\setcounter{proposition}{5}
\begin{proposition}[Adapted from Theorem 4 in \cite{GonzalezDiaz2019TwohiddenlayerFN}]
\label{prop:simplimap}
    Let us consider a simplicial map $\phi_c: |K|\to |L|$ between the underlying space of two finite pure simplicial complexes $K$ and $L$. Then a two-hidden-layer feed-forward network $\gN_{\phi}$ such that $\phi_c(x)=\gN_{\phi}(x)$ for all $x\in|K|$ can be explicitly defined. The size of $N_f$ is $D+d+k(D+1)+l(d+1)$, where $D=\text{dim}(|K|)$ and $d=\text{dim}(|L|)$, $k$ and $l$ are the number of simplices in $K$ and $L$, respectively.
\end{proposition}

\begin{theorem}[Main Theorem]
    Let $\gM=\gM_1\sqcup\gM_0\subset\R^D$ be a $d$-dimensional manifold from the thickened $1$-manifold family. $\gM_1$ and $\gM_0$ are two disjoint sub-manifolds of $\gM$ representing two classes. The condition number of $\gM$ is $\frac{1}{\tau}$ and the total Betti number of $\gM_1$ is $\beta$. Given a training set $\{(\rvx_i, y_i)|\rvx_i\in\gM, y_i\in\{0,1\}\}_{i=1}^n$, where $\rvx_i$ are sampled i.i.d. from $\gM$ by a uniform distribution, and $y_i=I_{\gM_1}(\rvx_i)$. For any $\delta>0$, if inequality~(\ref{eq:n}) holds, then for any $\epsilon>0$, there exists a ReLU network $g$ with depth at most $O(\log\beta+d\log\frac{1}{\tau} + \log\log\frac{1}{\tau\delta})$ and size at most $O(\frac{d^2\beta^2}{\epsilon}+\tau^{-d^2/2}\log^{d/2}\frac{1}{\tau\delta}+D\tau^{-d}\log\frac{1}{\tau\delta})$, such that
    \begin{equation}
       P(R(g)\le\epsilon)>1-\delta,
    \end{equation}
    where $R(g)=\int_{\gM}(g-I_{\gM_1})^2\mu(\mathbf{x})d\mathbf{x}$ with any continuous distribution $\mu$.
\end{theorem}
\begin{proof}
    Since $\gM=\gM_1\sqcup\gM_0$ is from thickened $1$-manifold family, it has a topological representative $\gM'=\gM'_1\sqcup\gM'_0\subset \R^d$, where $\gM'_1$ and $\gM'_0$ are topological representatives of $\gM_1$ and $\gM_2$, respectively.

    The proof follows by first constructing simplicial approximations $|K|$ and $|L|$ of $\gM$ and $\gM'$, respectively. Then we represent a simplicial homeomorphism $\phi:|K|\to |L|$ by a neural network $N_{\phi}$, where $K$ is constructed from $\gM$ and $L$ from $\gM'$. Built on the top of this, a projection from $\gM$ to its simiplicial approximation $|K|$ is represented by another network $N_p$. The overall network can be constructed by $g=h\circ N_{\phi}\circ N_p$. Note that $h$ is the function to approximate $I_{\gM'_1}$, but the data after projection and homeomorphism is from $|L_1|$. There should be an error in this approximation. However, we will show that by using the true risk, having $|L_1|\subseteq\gM'_1$ will make sure $||I_{|L_1|}-I_{\gM'_1}||_{L_2(\mu')}=0$. We move ahead by first constructing the network $g$, and then bounding the approximation error.

    \myparagraph{Network Construction.} Given $\gM$ is a compact submanifold of $\R^D$ and $\rvx_i$ are sampled according to a uniform distribution, by Proposition~\ref{prop:samplecomplex}, for all $0<r<\tau/2$ and $n>\lambda_1(log(\lambda_2)+log(\frac{1}{\delta}))$ ($n\sim O(\tau^{-d}\log(1/\tau\delta))$), $U=\bigcup_{i}B_{r}^D(\rvx_i)$ has the same homology as $\gM$ with probability higher than $1-\delta$. Note that every $B_{r}^D(\rvx_i)$ is contractible because $r\le\tau$. Therefore by the nerve theorem \citep{edelsbrunner2022computational}, the nerve of $U$ is homotopy equivalent to $\gM$. Note that $U$ is a collection of $\epsilon$-balls. The nerve of $U$ is the \v{C}ech complex, which is an abstract complex constructed as $\text{\v{C}ech}(r)=\{\sigma\subseteq X| \bigcap_{\rvx\in\sigma}B_{r}(\rvx)\neq 0\}$. But since the dimension of $\gM$ is $d$, it suffices to only consider simplices with dimension $\le d$. Delaunay complex is such a geometric construction that limits the dimension of simplices we get from a nerve. And in the other hand, we also do not want to lose the radius constraint. Here we construct the Alpha complex, a sub-complex of the Delaunay complex. It is constructed by intersecting each ball with the corresponding Voronoi cell, $R_{\rvx}(r)=B_{r}(\rvx)\cap V_{\rvx}$. The alpha complex is defined by
    \begin{equation}
        \text{Alpha}(r) = \{\sigma\subseteq X| \bigcap_{\rvx\in\sigma}R_{\rvx}(r)\neq 0\}.
    \end{equation}
    Based on the construction, $\text{Alpha}(r)$ also has the same homotopy type as $U$. \cite{Bern1995DihedralBF} provided the number of simplices in a Delaunay complex of $n$ vertices is bounded by $O(n^{\lceil d/2 \rceil})$. Since the Alpha complex is a sub-complex of Delaunay complex, the number of simplices in $\text{Alpha}(r)$ is also bounded by
    \begin{equation}
        O(n^{\lceil d/2 \rceil})=O(\tau^{-d^2/2}\log^{d/2}\frac{1}{\tau\delta})
    \end{equation}
    Denote $K=\text{Alpha}(r)$. 
    
    We claim that there exists a a vertex map $\phi:\rvx_i\to \rvx'_i$ for $i=1,...,n$, such that with probability higher than $1-\delta$, $U'=\bigcup_{i}B_{r'}^d(\rvx_i')$ has the same homology of $\gM$. We prove this claim after the proof.
    We can construct an alpha complex from $\{\rvx'_i\}_{i=1}^{n}$ in a similar way, $L=\text{Alpha}(r)$. The number of simiplices is also bounded by $O(\tau^{-d^2/2}\log^{d/2}\frac{1}{\tau\delta})$. 
    
    $\phi$ can be extended to a simiplicial map $\phi:|K|\to |L|$ by
    \begin{equation}
        \phi(\rvx) = \sum_{i=1}^n b_i(\rvx)\phi(\rvx_i).
    \end{equation}
    The map $b_i: |K|\to\R$ maps each point to its $i$-th barycentric coordinate. According to Proposition~\ref{prop:simplimap}, there exists a ReLU network $N_{\phi}$ with depth $4$ and size $O(\tau^{-d^2/2}\log^{d/2}\frac{1}{\tau\delta})$, such that $\phi(\rvx)=N_{\phi}(\rvx)$ for all $\rvx\in |K|$.
    
    Next we construct a network $N_p$ that projects $\gM$ to its simplicial approximation $|K|$. The point is projecting $\rvx\in\gM$ to its closest simplex $\sigma_{\rvx}$. According to the proof of theorem 3 in \cite{chartae2019stefan}, such projection can be represented as a neural network $N_p$ with depth at most $\log n +1$ and size at most $O(nD)$. Lastly, by Theorem~\ref{thm:topocomplex}, a neural network $h$ with depth at most $O(\log\beta)$ and size at most $O(\frac{d^2\beta^2}{\epsilon_1})$ can approximate $I_{\gM'_1}$ with error $R(h)\le\epsilon_1$. And by Lemma~\ref{lemma:func-comp}, $g=h\circ N_{\phi}\circ N_{p}$ has depth at most $O(\log(n\beta))$ and size $O(\frac{d^2\beta^2}{\epsilon_1}+\tau^{-d^2/2}\log^{d/2}\frac{1}{\tau\delta}+nD)$. Given $n\sim O(\tau^{-d}\log(1/\tau\delta))$, $g$ has depth at most $O(\log\beta+d\log\frac{1}{\tau} + \log\log\frac{1}{\tau\delta})$ and size at most $O(\frac{d^2\beta^2}{\epsilon_1}+\tau^{-d^2/2}\log^{d/2}\frac{1}{\tau\delta}+D\tau^{-d}\log\frac{1}{\tau\delta})$. Note that the probability of the existence for such network is larger than $(1-\delta)^2=1-2\delta+\delta^2>1-2\delta$. We let $\delta=2\delta$, such that with probability larger than $1-\delta$, neural network $g$ exists and $g$ has depth at most $O(\log\beta+d\log\frac{1}{\tau} + \log\log\frac{1}{\tau\delta})$ and size at most $O(\frac{d^2\beta^2}{\epsilon_1}+\tau^{-d^2/2}\log^{d/2}\frac{1}{\tau\delta}+D\tau^{-d}\log\frac{1}{\tau\delta})$.

    \myparagraph{Bounding Approximation Error.} Now it is time to bound the approximation error $R(g)$. We split $R(g)$ into two parts.
    \begin{align}
    \begin{split}
        R(g) &= ||g-I_{\gM_1}||_{L_2(\mu)} \\
        &\le ||h\circ N_{\phi}\circ N_{p}-I_{\gM'_1}\circ \phi \circ N_p||_{L_2(\mu)} + ||I_{\gM'_1}\circ \phi \circ N_p-I_{\gM_1}||_{L_2(\mu)} \\
        &= I_1 + I_2.
    \end{split}
    \end{align}
     We first show that $I_2=0$. Note that $N_p:\gM\to |K|$, and $\phi:|K|\to|L|$. We claim that for $\rvx\in\gM_1$, $\phi\circ N_p (\rvx)\in L_1$ and if $\rvx\in\gM_0$, $\phi\circ N_p (\rvx)\in L_0$. This is true because $\gM_1$ and $\gM_2$ are disjoint and $L$ is homotopy equivalent to $\gM$. Consequently, $I_{L_1}\circ \phi \circ N_p = I_{\gM_1}$. Given
     \begin{align}
     \begin{split}
         ||I_{\gM'_1}\circ \phi \circ N_p-I_{\gM_1}||_{L_2(\mu)} &= ||I_{\gM'_1}\circ \phi \circ N_p-I_{L_1}\circ \phi \circ N_p||_{L_2(\mu)} \\
         & = ||I_{L_1}-I_{\gM'_1}||_{L_2(\mu')}.
     \end{split}
     \end{align}
     The second equation is derived by letting $\mu'(\rvx') = \mu\circ\phi \circ N_p (\rvx)$.
     Now it suffices to show $||I_{L_1}-I_{\gM'_1}||_{L_2(\mu')}=0$. Note that $\mu'$ is a distribution supported on $|L|$, it can be naturally extended to $\gM'$ by set $\mu'(\rvx')=0$ if $\rvx'\in\gM'$ but $\rvx'\notin |L|$. For $\rvx'\in |L_1|$ but $\rvx'\notin\gM'_1$, like shown in Figure~\ref{fig:manifold_ext}, we extend $\gM_1'$ so that $\rvx'\in\gM'_1$. And such extension always exists due to the construction of $|L_1|$. Note that $|L_1|$ is a alpha complex constructed from a $r'$-cover, in a way that there will be an edge if and only if two covering balls have intersection. Hence, for any edge $(\rvx'_i, \rvx'_j)\in |L_1|$, the length $l_{ij}$ of it satisfies $l_{ij}<2r'<\tau'$. And notice the radius of the inner circle should be at least $\tau'$. Otherwise, the reach will be less than $\tau'$. Consequently, the edge of $|L_1|$ is always smaller than the radius of the inner circle. Therefore, one can always choose a $\gM_1'$, so that $\forall\rvx'\in |L_1|$ but $\rvx'\in\gM'_1$.

     \begin{figure}
         \centering
         \includegraphics[width=\linewidth]{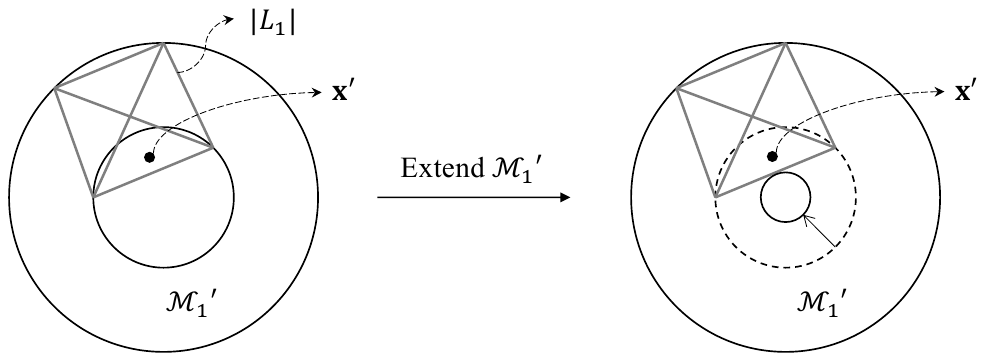}
         \caption{For a point $\rvx'\in |L_1|$ but $\rvx'\notin\gM'_1$, we extend $\gM'_1$ to make $\rvx'\in\gM'_1$. The extended $\gM'_1$ is still a topological representative.}
         \label{fig:manifold_ext}
     \end{figure}
    
    %Although this way $\mu'$ may not be continuous, Theorem~\ref{thm:topocomplex} still holds for it, because the claim, $\exists\delta$ such that $\int_{S_{\delta}}\mu'(\rvx')d\rvx'\le\epsilon$, still holds. 
    
    % The term $||I_{L_1}-I_{\gM'_1}||_{L_2(\mu')}$ is not likely to be zero because there are points , which will raise error. Note that points $\rvx'\in\gM'_1$ but $\rvx'\notin |L_1|$ will not cause any error because $\mu'(\rvx')=0$. However, the topological representative $\gM'$ is flexible in a way that we can adjust its radius. We claim that one can extent the boundary of $\gM'_1$, such that $|L_1|\subset\gM'_1$ and $\gM'_1$ is still the topological representative of $\gM_1$. We prove this in Claim~\ref{claim: extend}.
    
    %Since $|L_1|$ is homotopy equivalent to $\gM'_1$, $\gM'_1$ can be always extended in a way that it can include $|L_1|$. For example, for a arbitrary triangle, there are always an incircle and a circumcircle, and the region between them includes the entire triangle.
    %(Now if we extend the boundary of $\gM'_1$ by $r'$, as $\gM'_1=\gM'_1\cup(\bigcup_{\rvx'\in\partial\gM'_1}B_{r'}(\rvx'))$, the new $\gM'=\gM'_1\cup\gM'_0$ will still have the same homology since $r'<\tau'/2$, which indicates homeomorphic under the assumption of solid manifolds.) \MY{tbc} 
    After the expansion, $|L_1|\subseteq\gM'_1$. As a conclusion, $||I_{L_1}-I_{\gM'_1}||_{L_2(\mu')}=0$. Hereby, we have proved $I_2=0$.

    Now we settle $I_1$. Given $N_{\phi}$ is an exact representation of the simplicial map $\phi$,
    \begin{equation}
        I_1 = ||h-I_{\gM'_1}||_{L_2(\mu')}.
    \end{equation}
    According to Theorem~\ref{thm:topocomplex}, $I_1\le\epsilon_1$. Combined together, we have
    \begin{equation}
        R(g) \le \epsilon_1.
    \end{equation}
    Note that this inequality holds only with probability larger than $1-\delta$ because that is the probability we successfully recover the homology of $\gM$ by the training set and construct a simplicial homeomorphism.
\end{proof}

\begin{claim}
    $\gM\in\R^D$ is a $d$-dimensional manifold from thickened $1$-manifold family. Suppose there exists a set $\{\rvx_i\in\gM\}_{i=1}^{n}$ and radius $r$, such that $U=\bigcup_{i}B_{r}^D(\rvx_i)$ is a cover of $\gM$ and has the same homology. Then there exists a $\gM'\in\R^d$ that is a topological representative of $\gM$. Denote the homeomorphism between them $f$. Then with probability larger than $1-\delta$, $U'=\bigcup_{i}B_{r'}^d(f(\rvx_i))$ has the same homology as $\gM$.
\end{claim}
\begin{proof}
We let 
\begin{equation}
    c(r,\tau, \gM) = \frac{vol(\gM)}{(cos^d\theta_1)vol(B_{r/4}^d)}\left(\log \frac{vol(\gM)}{(cos^d\theta_2)vol(B_{r/8}^d)}+\log\frac{1}{\delta} \right),
\end{equation}
where $\theta_1=\arcsin{\frac{r}{8\tau}}$, $\theta_2=\arcsin{\frac{r}{16\tau}}$ and $0<r<\tau/2$.
Given a set $\{f(\rvx_i)\}_{i=1}^n$, apply proposition~\ref{prop:samplecomplex} to $\gM'$. If 
\begin{equation}
    n>c(r', \tau', \gM'),
\end{equation}
then with probability $1-\delta$, $U'=\bigcup_{i}B_{r'}^d(f(\rvx_i))$ has the same homology as $\gM'$, with $r'<\tau'/2$.

Note that $n$ already satisfy that $n>c(r, \tau, \gM)$, it suffices to show $c(r, \tau, \gM)>c(r', \tau', \gM')$. Since $\gM'$ is one of topological representatives of $\gM$, we can always choose the radius of the fundamental members in $\gM'$ and choose the distance between $\gM'_1$ and $\gM'_2$, to make sure that $\tau'>\tau$ and $vol(\gM')<vol(\gM)$. Hence, we can choose $r$ and $r'$, such that $B_{r'}^d>B_{r}^d$. With the same $\delta$, we have proved that $c(r, \tau, \gM)>c(r', \tau', \gM')$.

\end{proof}

%% file: main-arxiv.bbl
\begin{thebibliography}{33}
\providecommand{\natexlab}[1]{#1}
\providecommand{\url}[1]{\texttt{#1}}
\expandafter\ifx\csname urlstyle\endcsname\relax
  \providecommand{\doi}[1]{doi: #1}\else
  \providecommand{\doi}{doi: \begingroup \urlstyle{rm}\Url}\fi

\bibitem[Amenta \& Bern(1998)Amenta and Bern]{amenta1998surface}
Nina Amenta and Marshall Bern.
\newblock Surface reconstruction by voronoi filtering.
\newblock In \emph{Proceedings of the fourteenth annual symposium on Computational geometry}, pp.\  39--48, 1998.

\bibitem[Arora et~al.(2018)Arora, Basu, Mianjy, and Mukherjee]{arora2018understanding}
Raman Arora, Amitabh Basu, Poorya Mianjy, and Anirbit Mukherjee.
\newblock Understanding deep neural networks with rectified linear units.
\newblock In \emph{International Conference on Learning Representations}, 2018.

\bibitem[Bern et~al.(1995)Bern, Chew, Eppstein, and Ruppert]{Bern1995DihedralBF}
Marshall~W. Bern, L.~Paul Chew, David Eppstein, and Jim Ruppert.
\newblock Dihedral bounds for mesh generation in high dimensions.
\newblock In \emph{ACM-SIAM Symposium on Discrete Algorithms}, 1995.

\bibitem[Bianchini \& Scarselli(2014)Bianchini and Scarselli]{complexityneural2014bianchini}
Monica Bianchini and Franco Scarselli.
\newblock On the complexity of neural network classifiers: A comparison between shallow and deep architectures.
\newblock \emph{IEEE Transactions on Neural Networks and Learning Systems}, 2014.

\bibitem[Bott et~al.(1982)Bott, Tu, et~al.]{bott1982differential}
Raoul Bott, Loring~W Tu, et~al.
\newblock \emph{Differential forms in algebraic topology}, volume~82.
\newblock Springer, 1982.

\bibitem[Buchanan et~al.(2021)Buchanan, Gilboa, and Wright]{buchanan2021deep}
Sam Buchanan, Dar Gilboa, and John Wright.
\newblock Deep networks and the multiple manifold problem.
\newblock In \emph{International Conference on Learning Representations}, 2021.

\bibitem[Chen et~al.(2019)Chen, Jiang, Liao, and Zhao]{lowmanifoldChen2019}
Minshuo Chen, Haoming Jiang, Wenjing Liao, and Tuo Zhao.
\newblock Efficient approximation of deep relu networks for functions on low dimensional manifolds.
\newblock In \emph{Advances in Neural Information Processing Systems}, 2019.

\bibitem[Cybenko(1989)]{cybenko1989approximation}
George Cybenko.
\newblock Approximation by superpositions of a sigmoidal function.
\newblock \emph{Mathematics of control, signals and systems}, 2\penalty0 (4):\penalty0 303--314, 1989.

\bibitem[Dikkala et~al.(2021)Dikkala, Kaplun, and Panigrahy]{lsh2021dikkala}
Nishanth Dikkala, Gal Kaplun, and Rina Panigrahy.
\newblock For manifold learning, deep neural networks can be locality sensitive hash functions.
\newblock \emph{arXiv:2103.06875}, 2021.

\bibitem[Edelsbrunner \& Harer(2022)Edelsbrunner and Harer]{edelsbrunner2022computational}
Herbert Edelsbrunner and John~L Harer.
\newblock \emph{Computational topology: an introduction}.
\newblock American Mathematical Society, 2022.

\bibitem[Eldan \& Shamir(2016)Eldan and Shamir]{eldan2016power}
Ronen Eldan and Ohad Shamir.
\newblock The power of depth for feedforward neural networks.
\newblock In \emph{29th Annual Conference on Learning Theory}, 2016.

\bibitem[Federer(1959)]{federer1959curvature}
Herbert Federer.
\newblock Curvature measures.
\newblock \emph{Transactions of the American Mathematical Society}, 93\penalty0 (3):\penalty0 418--491, 1959.

\bibitem[Gonzalez-Diaz et~al.(2019)Gonzalez-Diaz, Guti{\'e}rrez-Naranjo, and Paluzo-Hidalgo]{GonzalezDiaz2019TwohiddenlayerFN}
Rocio Gonzalez-Diaz, Miguel~Angel Guti{\'e}rrez-Naranjo, and Eduardo Paluzo-Hidalgo.
\newblock Two-hidden-layer feedforward neural networks are universal approximators: A constructive approach.
\newblock \emph{Neural networks : the official journal of the International Neural Network Society}, 2019.

\bibitem[Grigsby \& Lindsey(2022)Grigsby and Lindsey]{grigsby2022transversality}
J~Elisenda Grigsby and Kathryn Lindsey.
\newblock On transversality of bent hyperplane arrangements and the topological expressiveness of relu neural networks.
\newblock \emph{SIAM Journal on Applied Algebra and Geometry}, 2022.

\bibitem[Guss \& Salakhutdinov(2018)Guss and Salakhutdinov]{guss2018characterizing}
William~H Guss and Ruslan Salakhutdinov.
\newblock On characterizing the capacity of neural networks using algebraic topology.
\newblock \emph{arXiv preprint arXiv:1802.04443}, 2018.

\bibitem[Hanin(2017)]{hanin2017approximating}
Boris Hanin.
\newblock Approximating continuous functions by relu nets of minimal width.
\newblock \emph{arXiv preprint arXiv:1710.11278}, 2017.

\bibitem[Hanin \& Rolnick(2019)Hanin and Rolnick]{hanin2019complexity}
Boris Hanin and David Rolnick.
\newblock Complexity of linear regions in deep networks.
\newblock In \emph{International Conference on Machine Learning}, 2019.

\bibitem[Hatcher(2002)]{hatcher2002algebraic}
Allen Hatcher.
\newblock \emph{Algebraic Topology}.
\newblock Cambridge University Press, 2002.

\bibitem[Hornik(1989)]{hornik1989multilayer}
Kurt Hornik.
\newblock Multilayer feedforward networks are universal approximators.
\newblock \emph{Neural networks}, 2\penalty0 (5):\penalty0 359--366, 1989.

\bibitem[Jolliffe \& Cadima(2016)Jolliffe and Cadima]{pcaJolliffe2016}
Ian Jolliffe and Jorge Cadima.
\newblock Principal component analysis: A review and recent developments.
\newblock \emph{Philosophical Transactions of the Royal Society A: Mathematical, Physical and Engineering Sciences}, 374:\penalty0 20150202, 04 2016.

\bibitem[Leshno et~al.(1993)Leshno, Lin, Pinkus, and Schocken]{leshno1993multilayer}
Moshe Leshno, Vladimir~Ya Lin, Allan Pinkus, and Shimon Schocken.
\newblock Multilayer feedforward networks with a nonpolynomial activation function can approximate any function.
\newblock \emph{Neural networks}, 6\penalty0 (6):\penalty0 861--867, 1993.

\bibitem[Munkres(2018)]{munkres2018elements}
James~R Munkres.
\newblock \emph{Elements of algebraic topology}.
\newblock CRC press, 2018.

\bibitem[Naitzat et~al.(2020)Naitzat, Zhitnikov, and Lim]{naitzat2020topology}
Gregory Naitzat, Andrey Zhitnikov, and Lek-Heng Lim.
\newblock Topology of deep neural networks.
\newblock \emph{The Journal of Machine Learning Research}, 21\penalty0 (1):\penalty0 7503--7542, 2020.

\bibitem[Narayanan \& Mitter(2010)Narayanan and Mitter]{samplecomplexNaray2010}
Hariharan Narayanan and Sanjoy Mitter.
\newblock Sample complexity of testing the manifold hypothesis.
\newblock In \emph{Advances in Neural Information Processing Systems}, 2010.

\bibitem[Narayanan \& Niyogi(2009)Narayanan and Niyogi]{narayanan2009sample}
Hariharan Narayanan and Partha Niyogi.
\newblock On the sample complexity of learning smooth cuts on a manifold.
\newblock In \emph{COLT}, 2009.

\bibitem[Niyogi et~al.(2008)Niyogi, Smale, and Weinberger]{Niyogi2008Homology}
Partha Niyogi, Stephen Smale, and Shmuel Weinberger.
\newblock Finding the homology of submanifolds with high confidence from random samples.
\newblock \emph{Discrete \& Computational Geometry}, 2008.

\bibitem[Ramamurthy et~al.(2019)Ramamurthy, Varshney, and Mody]{labeledcomplex2019ramamurthy}
Karthikeyan~Natesan Ramamurthy, Kush Varshney, and Krishnan Mody.
\newblock Topological data analysis of decision boundaries with application to model selection.
\newblock In \emph{Proceedings of International Conference on Machine Learning}, 2019.

\bibitem[Rieck et~al.(2019)Rieck, Togninalli, Bock, Moor, Horn, Gumbsch, and Borgwardt]{rieckneural}
Bastian Rieck, Matteo Togninalli, Christian Bock, Michael Moor, Max Horn, Thomas Gumbsch, and Karsten Borgwardt.
\newblock Neural persistence: A complexity measure for deep neural networks using algebraic topology.
\newblock In \emph{International Conference on Learning Representations}, 2019.

\bibitem[Roweis \& Saul(2000)Roweis and Saul]{lleRoweis2000}
Sam~T. Roweis and Lawrence~K. Saul.
\newblock Nonlinear dimensionality reduction by locally linear embedding.
\newblock \emph{Science}, 290\penalty0 (5500):\penalty0 2323--2326, 2000.

\bibitem[Safran \& Shamir(2016)Safran and Shamir]{Safran2016DepthWidthTI}
Itay Safran and Ohad Shamir.
\newblock Depth-width tradeoffs in approximating natural functions with neural networks.
\newblock In \emph{International Conference on Machine Learning}, 2016.

\bibitem[Schmidt-Hieber(2019)]{schmidt2019deep}
Johannes Schmidt-Hieber.
\newblock Deep relu network approximation of functions on a manifold.
\newblock \emph{arXiv preprint arXiv:1908.00695}, 2019.

\bibitem[Schonsheck et~al.(2019)Schonsheck, Chen, and Lai]{chartae2019stefan}
Stefan~C. Schonsheck, Jie Chen, and Rongjie Lai.
\newblock Chart auto-encoders for manifold structured data.
\newblock \emph{CoRR}, abs/1912.10094, 2019.

\bibitem[van~der Maaten \& Hinton(2008)van~der Maaten and Hinton]{tsneVandermaaten2008}
Laurens van~der Maaten and Geoffrey Hinton.
\newblock Visualizing data using t-sne.
\newblock \emph{Journal of Machine Learning Research}, 9\penalty0 (86):\penalty0 2579--2605, 2008.

\end{thebibliography}
